\documentclass[preprint,3p,5pt,authoryear]{elsarticle}

\usepackage{amssymb}
\usepackage{float}
\usepackage{amsmath}
\usepackage{algorithm}
\usepackage{algorithmic}
\usepackage{multirow}
\usepackage{longtable}
\usepackage{graphicx}
\usepackage{makecell}
\usepackage{xcolor}
\usepackage[colorlinks=true, allcolors=black]{hyperref}
\renewcommand{\textcolor}[2]{#2}

\journal{Computers \& Operations Research}

\begin{document}

\begin{frontmatter}



\title{TuneNSearch: a hybrid transfer learning and local search approach for solving vehicle routing problems} 


\author[UCAddress]{Arthur Corrêa}
\author[UCAddress]{Cristóvão Silva}
\author[SCAILAddress]{Liming Xu}
\author[SCAILAddress]{Alexandra Brintrup}
\author[UCAddress]{Samuel Moniz\corref{cor1}}

\cortext[cor1]{Corresponding author.\\
Email: samuel.moniz@dem.uc.pt\\
Address: Pólo II da Universidade de Coimbra, Rua Luís Reis Santos, 3030-788, Coimbra, Portugal\\
Phone: +351 239 790 712}

\affiliation[UCAddress]{organization={Department of Mechanical Engineering, CEMMPRE, ARISE, Universidade de Coimbra}, 
	city={Coimbra},
	country={Portugal}}

\affiliation[SCAILAddress]{organization={Supply Chain AI Lab, Institute for Manufacturing, Department of Engineering, University of Cambridge}, 
	city={Cambridge},
	postcode={CB3 0FS}, 
	country={UK}}

\begin{abstract}
This paper introduces TuneNSearch, a hybrid transfer learning and local search approach for addressing diverse variants of the vehicle routing problem (VRP). Our method uses reinforcement learning to generate high-quality solutions, which are subsequently refined by an efficient local search procedure.  To ensure broad adaptability across VRP variants, TuneNSearch begins with a pre-training phase on the multi-depot VRP (MDVRP), followed by a fine-tuning phase to adapt it to other problem formulations. The learning phase utilizes a Transformer-based architecture enhanced with edge-aware attention, which integrates edge distances directly into the attention mechanism to better capture spatial relationships inherent to routing problems. We show that the pre-trained model generalizes effectively to single-depot variants, achieving performance comparable to models trained specifically on single-depot instances. \textcolor{blue}{Simultaneously, it maintains strong performance   on multi-depot variants, an ability that models pre-trained solely on single-depot problems lack. For example, on 100-node instances of multi-depot variants, TuneNSearch outperforms a model pre-trained on the CVRP by 44\%. In contrast, on 100-node instances of single-depot variants, TuneNSearch performs similar to the CVRP model. To validate the effectiveness of our method, we conduct extensive computational experiments on public benchmark and randomly generated instances. Across multiple CVRPLIB and TSPLIB datasets, TuneNSearch consistently achieves performance deviations of less than 3\% from the best-known solutions in literature, compared to 6–25\% for other neural-based models, depending on problem complexity. Overall, our approach demonstrates strong generalization to different problem sizes, instance distributions, and VRP formulations, while maintaining polynomial runtime complexity despite the integration of the local search algorithm.}
\end{abstract}



\end{frontmatter}



\section{Introduction}
\label{sec:intro}

Vehicle routing problems (VRP) are a class of combinatorial optimization problems that hold particular importance in both academic literature and real-world settings. These problems are particularly relevant in fields such as city and food logistics, transportation and drone delivery  \citep{Cattaruzza2017,Li2019,Wang2019,Wu2023}. In short, they entail finding the most efficient route to visit a set of predefined nodes (such as delivery or service locations) using one or more vehicles, with the goal of minimizing the total traveled distance. Routing problems encompass numerous variants, including the traveling salesman problem (TSP), the capacitated vehicle routing problem (CVRP), the multi-depot vehicle routing problem (MDVRP), CVRP with backhauls, and other variants that introduce additional constraints \citep{Elatar2023}.

The combinatorial characteristics of the VRP and its variants make these problems very difficult to solve optimally. Their inherent NP-hardness and intractability often render exact methods impractical, especially when solving large-scale problems with complicated constraints. Alternatively, meta-heuristics rely on search techniques to explore the solution space more efficiently. \textcolor{blue}{Historically, meta-heuristics can be classified into single-solution based and population-based methods \citep{Laporte2009}. Single-solution based methods start from an individual solution and iteratively improve it through local search processes. They are relatively simple to implement and exhibit strong intensification, since they focus on promising areas of the solution search space. Examples of single-solution based methods include tabu search \citep{Gendreau1994}, simulated annealing \citep{Osman1993}, variable neighborhood search \citep{Mladenovic1997} and adaptive large neighborhood search \citep{Ropke2006}. Their main limitation, however, is that they have a high risk of getting trapped in local minimum regions. In this regard, population-based algorithms show better exploration and diversification, since they maintain multiple solutions simultaneously, that is, a population of solutions. Common population-based methods include genetic algorithms \citep{Baker2003}, ant colony optimization \citep{Dorigo1997} and particle swarm optimization \citep{Ai2009}.}

\textcolor{blue}{To leverage the strengths of both approaches, many state-of-the-art VRP methods combine single-solution and population-based strategies. One of the most prominent examples is the hybrid genetic search \citep{Vidal2022}, which integrates a local search method at each iteration of a genetic algorithm to improve the quality of offspring solutions. It has achieved state-of-the-art performance, outperforming multiple other well-known methods. More recently, \citet{Wouda2024} proposed PyVRP, a computationally efficient open-source implementation of the hybrid genetic search, designed to handle a range of VRP variants, instead of being tied to a single problem type.} Despite presenting good results when solving complicated problems, compared with exact approaches, even the most efficient meta-heuristics still face significant computational burden as problem size grows. As a result, their applicability is limited in real-world cases, where rapid and precise decision-making is critical \citep{Li2023}.

In recent years, approaches based on neural networks have been gaining traction as an alternative to exact methods and meta-heuristics. These methods use neural networks to learn policies that can approximate good quality solutions with minimal computational overhead and little domain-specific knowledge. Most neural-based methods are trained using supervised or reinforcement learning algorithms and can be divided into two categories: construction and improvement approaches. The former is more predominant and refers to algorithms that can generate solutions in an end-to-end fashion, such as \citet{Kool2019}, \citet{Kwon2020} and \citet{Zhou2023}. The latter methods have the capability to learn policies that iteratively improve an initially generated solution \citep{Hudson2022,Roberto2020,Wu2022,Xin2021}.

While neural-based methods demonstrate promising results, they are often tailored for specific types of routing problems, similar to meta-heuristics, which are typically designed for particular cases. This specialization limits their ability to generalize across different problem variants. Recent research efforts have sought to address this challenge. For instance, \citet{Liu2024} proposed a multi-task learning method that utilizes attribute composition to solve different VRP variants. Meanwhile, \citet{Zhou2024}, explored a different approach, using a mixture-of-experts model to enhance cross-task generalization. Although these methods improve flexibility, they typically require extensive training (up to 100 million instances, or 5+ days of training time, depending on the hardware). Moreover, their performance often falls short when compared to models trained specifically for individual variants, especially on more complex tasks. In contrast, \citet{Lin2024} explored a fine-tuning approach that adapts a backbone model, pre-trained on a standard TSP, to effectively solve other variants. While this method outperforms models trained independently for each variant, the fine-tuning phase demands a computational effort and training duration comparable to training a new model from scratch. Furthermore, existing multi-task and transfer learning approaches tend to focus exclusively on single-depot variants, overlooking the MDVRP, which more accurately reflects real-world logistics applications. 

From a practical perspective, the ability to develop a model that can generalize across different problems has profound implications, particularly in the most challenging manufacturing settings. As the demand for highly adaptable decision-support frameworks increases, practitioners are looking for solutions that can be customized to meet their specific needs \citep{Jan2023}. By reducing the reliance on multiple specialized models, businesses can conserve resources (both computational and manpower resources) and minimize the time required to develop and deploy solutions across different scenarios. This versatility could offer significant operational advantages, enabling companies to adapt more easily to changing routing conditions or new problem variants. However, as evidenced by the trade-offs in existing methods, achieving a balance between computational efficiency and solution quality remains a crucial challenge.

Motivated by this challenge, we introduce TuneNSearch, a hybrid framework that combines transfer learning with an efficient local search procedure to solve different VRP variants. Transfer learning is a machine learning technique that allows a model trained on one task to be adapted for a different but related task \citep{Pan2010}. In this way, previously acquired knowledge can be leveraged to improve the learning performance and reduce training time. Specifically, we propose pre-training our model on the MDVRP, a rich and challenging problem setting that has been underexplored in neural approaches. \textcolor{blue}{By exploiting the complexity inherent to the MDVRP, our goal is for the model to adapt well to both single-depot and multi-depot scenarios, a capability missing in models pre-trained exclusively on single-depot data.} Given that existing neural-based methods often struggle with generalization, particularly for larger instances, our approach features a hybridization combining machine learning with a high-performance local search algorithm. Therefore, a two-stage process is proposed, where an efficient local search refines solutions after the initial inference stage. \textcolor{blue}{As a result, TuneNSearch achieves significant improvements in generalization, substantially reducing the performance difference compared to existing methods, such as \citet{Kwon2020}, \citet{Li2024} and \citet{Zhou2024}.} The key contributions of this work include:

\begin{itemize}
	\item Improvement of the inference process by integrating an efficient local search method. This local search algorithm employs a set of different operators to improve the solutions obtained by the reinforcement learning model, resulting in significant performance gains with small additional computational cost. As a result, the proposed method achieves new state-of-the-art performance among neural-based approaches.
	
	\item Development of a transfer learning method that begins with pre-training on the MDVRP, designed to adapt effectively to various VRP variants. Unlike simpler variants, the MDVRP incorporates a multi-depot structure in which vehicles can depart from multiple locations, thereby allowing the model to capture richer and more complex representations. Computational results show that this approach facilitates a more efficient generalization to solve both single- and multi-depot VRP variants.
	
	\item Proposal of a novel learning framework to solve the MDVRP. Our method employs a Transformer architecture \citep{Vaswani2017} using the policy optimization with multiple optima (POMO) \citep{Kwon2020}. To better capture the complex spatial and relational structure of MDVRP instances, we integrate a residual edge-aware graph attention network (E-GAT), inspired by \citet{Lei2022}, but uniquely adapted to work within the POMO framework and extended to the multi-depot setting. This allows the model to incorporate edge distance information directly into the attention mechanism, leading to more accurate representations.  Our method achieves substantial empirical improvements, outperforming the multi-depot multi-type attention (MD-MTA) approach proposed by \citet{Li2024}. 
	\textcolor{blue}{Specifically, in 50-node instances, we reduced the relative deviation from the best-performing method in our experiments by a factor of 6, and on 100-node instances by a factor of more than 2.}
	
	\item To verify the effectiveness of TuneNSearch, we solve numerous large-scale datasets from CVRPLIB and TSPLIB. \textcolor{blue}{On average, the solutions for each dataset deviate by no more than 3\% from the best-known solutions in literature. We also assessed the performance of our method on thousands of randomly generated instances. For 100-node instances, across all VRP variants, it obtains an average relative performance deviation of less than 4\% compared to PyVRP’s hybrid genetic search \citep{Wouda2024}, while requiring only a fraction of the computational time.  Overall, TuneNSearch demonstrates robust cross-distribution, cross-size and cross-task generalization, consistently exceeding other neural-based methods.}
\end{itemize}

The rest of this paper is organized as follows. Section~\ref{sec:review} discusses the relevant literature. Section~\ref{sec:prelim} provides important preliminaries for this work. Section~\ref{sec:method} elaborates on the model architecture, including the encoder, decoder, fine-tuning process and local search mechanism. Section~\ref{sec:experiments} displays the computational experiments performed. Finally, Section~\ref{sec:conclusion} draws conclusions and limitations, and envisions possible future work. 

\section{Related work}
\label{sec:review}

This section briefly reviews three main research areas pertinent to this study. First, it examines optimization approaches for solving VRPs, including exact methods and meta-heuristics. Second, it discusses the most relevant neural-based methods developed in this field. Third, it explores multi-task learning and transfer learning techniques, which is an emerging field dedicated to creating models that can effectively address various VRP variants.

\subsection{Solving VRPs with exact methods and meta-heuristics}
\label{sec:2.1}

Over the last few decades, a variety of methods have been proposed to address routing problems, including exact methods and meta-heuristic algorithms. Exact methods are fundamentally limited in their ability to guarantee optimal solutions within polynomial time. As a result, their application is typically limited to small- and medium-sized problem instances. The development and application of exact methods for routing problems has been reviewed by \citet{Baldacci2012} and \citet{Zhang2022VRP}. Most of these methods involve techniques such as branch-and-cut, dynamic programming and set partitioning formulations. More recently, \citet{Pessoa2020} introduced a generic solver based on a branch-cut-and-price algorithm capable of solving different routing problem variants. Nevertheless, the proposed method remains computationally intractable for instances involving more than a few hundred nodes.

Alternatively, meta-heuristic algorithms are far more prevalent than exact methods in the literature \citep{Braekers2016}. Unlike exact approaches, meta-heuristics do not guarantee optimal solutions, as a complete search of the solution space cannot be proven. \textcolor{blue}{However, these methods tend to be more efficient, utilizing advanced exploration techniques to find high-quality solutions, and often even reaching optimal ones in significantly less time. \citet{Gendreau1994} proposed TABUROUTE, one of the first effective tabu search methods for routing problems. Specifically, the authors addressed the CVRP under route distance limit constraints. During the search process, they allow temporary violations of constraints, managing them with dynamically adjusted penalties in the objective function, improving exploration compared to methods that stay strictly within feasible regions of the search space.} Subsequently, \citet{Renaud1996} presented one of the most influential meta-heuristic methods for the MDVRP, designing a tabu search algorithm that constructs an initial solution by assigning each customer to its nearest depot. The approach consists of three phases: fast-improvement, intensification and diversification. The first phase aims at rapidly improving the incumbent solution using different search operators. The intensification phase then focuses on exploring high-potential areas of the search space, while the last one introduces controlled perturbations to escape local minimum and enable broader exploration. \textcolor{blue}{\citet{Mladenovic1997} later introduced the variable neighborhood search meta-heuristic, a method that systematically changes neighborhood structures during the search process. Each time the algorithm explores a different neighborhood of the incumbent solution, it applies a local search to find a local optimum within that neighborhood. This local search serves to exploit the current region of the solution space, while the systematic change of neighborhoods allows the algorithm to escape local minimum. Although initially tested on the TSP, the method’s effectiveness have established it as a standard within combinatorial optimization literature. After, \citet{Baker2003} proposed one of the first genetic algorithms for the classic VRP, offering a new perspective on solving routing problems. The authors also developed a hybrid method that incorporates neighborhood search mechanisms within the genetic algorithm to improve the balance between exploration and intensification. This hybridization of single-solution based and population-based methods would later become a widely established practice in VRP research. Computational results demonstrated the effectiveness of the genetic algorithm, showing performance competitive with other state-of-the-art methods at the time, including a tabu search and a simulated annealing algorithm. \citet{Ai2009} presented a particle swarm optimization algorithm for the VRP with simultaneous delivery and pickup. Tests on multiple benchmark datasets showed that their method performed competitively against other algorithms and even achieved new best-known solutions  for various benchmark instances. Although not directly based on the particle swarm optimization method of \citet{Ai2009}, \citet{Marinakis2010} explored a related idea by hybridizing a genetic algorithm with particle swarm optimization for the CVRP, allowing individual solutions within the population to evolve.} This approach reduced computational time and improved scalability to larger problem instances compared to other meta-heuristics.

\citet{Helsgaun2017} introduced LKH-3, a heuristic method capable of solving various routing problem variants. It transforms the problems into a constrained TSP and utilizes the LKH local search from \citet{Helsgaun2000} to effectively explore the solution space. \citet{Lopes2019} designed a multi-agent meta-heuristic framework combined with reinforcement learning for solving the VRP with time windows. In this framework, each meta-heuristic is represented as an autonomous agent that collaborates with others, enhancing solution quality through cooperative behavior. \citet{Vidal2022} presented a hybrid genetic search for the CVRP and introduced the new \textit{Swap*} operator. Rather than swapping two customers directly in place, this operator proposes exchanging two customers from different routes by inserting them into any position on the opposite route. \textcolor{blue}{By combining elements from both single-solution based and population-based meta-heuristics, this work has become a standard in CVRP literature, achieving state-of-the-art performance across various benchmarks.} \citet{Kalatzantonakis2023} presented a hybrid approach between reinforcement learning and a variable neighborhood search for the CVRP, utilizing different upper confidence bound algorithms for adaptive neighborhood selection. More recently, \citet{Wouda2024} introduced PyVRP, an open-source VRP solver package. PyVRP offers a high-performance implementation of the hybrid genetic search algorithm \citep{Vidal2022} and supports extensive customization options, making it suitable for a variety of VRP variants. Lastly, OR-Tools \citep{Furnon2024} is a general-purpose optimization toolkit designed to solve a wide range of combinatorial problems. It includes a robust routing library capable of addressing various VRP variants, offering greater versatility than the previously cited works. OR-Tools uses a first-solution heuristic to generate an initial solution, followed by a guided local search to iteratively improve it. Subsequently, it employs a constraint programming approach to check whether the solution satisfies all specified constraints.

\subsection{Neural-based combinatorial optimization for VRPs}
\label{sec:2.2}

Neural-based methods surged in recent years as an alternative to solve combinatorial problems \citep{Bengio2021,Mazyavkina2021}. By recognizing patterns in data, these methods can learn policies, obtaining high-quality solutions in polynomial time, even for large and hard to solve instances.  

As previously discussed, neural-based approaches can be categorized into construction or improvement methods. In what follows, we focus first on construction-based methods, which learn policies to incrementally build a solution. \citet{Vinyals2015} introduced Pointer Networks, a sequence-to-sequence model that addressed the problem of variable-sized outputs by using a ‘pointer’ mechanism to select elements from the input sequence as the output. Their approach was applied to solve the TSP and trained using supervised learning, being an early demonstration of the potential of neural networks for combinatorial optimization. Later, \citet{Bello2017} built on this approach by using a similar model architecture but training it with reinforcement learning. This eliminated the need for (near)-optimal labels and led to improved performance over Pointer Networks. \citet{Nazari2018} extended the architecture of Pointer Networks to handle dynamic elements in problems. Their approach proved effective in solving more challenging combinatorial problems, such as the stochastic VRP and VRP with split deliveries. \citet{Kool2019} made a significant contribution to recent literature by proposing an attention model that utilizes a Transformer architecture \citep{Vaswani2017}. Trained using reinforcement learning, their method outperformed previous methods across a variety of combinatorial problems, representing a major advancement in the field. \citet{Kwon2020} introduced POMO, a reinforcement learning approach which draws on  solution symmetries to improve results when compared to the attention model. \textcolor{blue}{POMO also introduced an instance augmentation technique which reformulates a given problem by applying transformations, such as flipping or rotating the Euclidean map of node coordinates, to generate alternative instances that lead to the same solution.} This technique forces the exploration of a wider range of potential solutions, enhancing model performance during inference. These works can be considered the backbone of the published research on routing problems, providing inspiration for a wide variety of subsequent studies \citep{Bi2025,Chalumeau2023,Fitzpatrick2024,Grinsztajn2023,Kim2022,Kwon2021,Lei2022,Lin2024, Luo2023,Pirnay2024,Xin2020,Zhou2023,Zhou2024b}.

Alternatively, improvement-based methods focus on learning to iteratively refine an initial solution through a structured search process, often drawing inspiration from traditional local search or large neighborhood search algorithms. While these methods are far less prevalent than construction-based approaches, they typically produce higher-quality solutions. However, this comes at the cost of significantly increased inference time. One of the first such approaches in VRP literature, NeuRewriter, was introduced by \citet{Chen2019}. Their approach, trained via reinforcement learning, learns region-picking and rule-picking policies to improve an initially generated solution until convergence. Later, \citet{Hottung2020} proposed incorporating a large neighborhood search as the foundation for the search process. They manually designed two destroy operators, while a deep neural network guided the repair process. \citet{Ma2021} introduced a dual-aspect collaborative Transformer, which learns separate embeddings for node and positional features. Their model also featured a cyclic encoding technique, which captures the symmetry of VRP problems to enhance generalization. \citet{Wu2022} developed a Transformer-based model for solving the TSP and CVRP, which parameterizes a policy to guide the selection of the next solution by integrating 2-opt and swap operators. These studies laid the foundation for many other works, influencing further advancements on the field \citep{Hudson2022,Kim2021,Ma2023,Roberto2020}.

\subsection{Multi-task learning and transfer learning for VRPs}
\label{sec:2.3}

Most algorithms, whether neural-based or not, are restricted to addressing specific VRP variants. Some machine learning techniques offer greater versatility, enabling the development of models that are not bound to a single task. Among these techniques, multi-task learning and transfer learning hold particular prominence \citep{Pan2010,Zhang2022}. Multi-task learning involves training a model simultaneously on data from multiple related but distinct tasks. In this manner, the model can effectively learn shared features and representations across various tasks, improving its generalization ability. In contrast, transfer learning focuses on pre-training a model on a single task and subsequently adapting it to a specific task. This is achieved by loading the pre-trained model’s parameters and making minor adjustments to it, which is faster and more efficient than training a model from the beginning.

While these techniques have been widely studied in computer vision \citep{Yuan2012} and natural language processing \citep{Dong2019}, their applications in combinatorial optimization remain relatively new. Recently, a few recent studies have begun exploring these methods in this domain, all utilizing reinforcement learning for training. \citet{Lin2024} proposed pre-training a backbone model on a standard TSP and subsequently fine-tuning it to adapt to other routing variants, including the orienteering problem, the prize collecting TSP and CVRP. Their approach modified the neural network architecture of the pre-trained model by incorporating additional layers tailored to the unique constraints of each routing variant considered in the fine-tuning phase. \citet{Liu2024} modified the attention model to include an attribute composition block. This technique updates a problem-specific attribute vector, which dynamically activates or deactivates relevant problem features depending on the VRP variant being solved. The model includes four attributes that represent capacity constraints, open routes, time windows, and route limits. Essentially, it functions as a multi-task learning model trained on data from various VRP variants. \citet{Zhou2024} aimed to improve generalization by incorporating a mixture-of-experts layer and a gating network. Specifically, the mixture-of-experts consists of multiple specialized sub-models, or “experts”, each one designed to handle different problem variants. The gating network then selects which experts to activate depending on the input, enabling the model to generalize to various tasks more effectively.

While these approaches improve flexibility and generalization, they still face some limitations. First, they often require substantial training resources (on the order of several days of compute time) and still fall short of the solution quality offered by specialized models or methods such as OR-Tools. Second, their generalization ability is often limited to single-depot VRP variants, which do not capture the operational complexity of more realistic logistics settings. The MDVRP, for example, is essential for modeling real-world logistics systems that involve multiple dispatch centers, such as urban distribution networks or decentralized supply chains, but remains underexplored in the context of neural combinatorial optimization.

\textcolor{blue}{In response, we propose TuneNSearch, a new transfer learning method initially pre-trained on the MDVRP and fine-tuned for multiple VRP variants.} Our approach distinguishes itself in three key ways: First, to improve the encoding of VRP’s features, we integrate POMO with the E-GAT encoder. While the residual E-GAT model has shown improvements over the attention model \citep{Lei2022}, to the best of our knowledge, no prior work has combined it with POMO. We demonstrate that this combination enables a more effective encoding, making a better use of the multiple starting nodes introduced by POMO; \textcolor{blue}{Second, while we draw inspiration from \citet{Lin2024}, we propose pre-training our model on the MDVRP, a significantly more complex problem than the TSP.} This allows the model to learn richer features, facilitating a more effective knowledge transfer across different VRP variants. \textcolor{blue}{On single-depot problems, TuneNSearch matches the performance of models trained on the CVRP, while performing substantially better on MDVRP variants;} Third, most existing neural-based approaches rely solely on machine learning, with little integration with optimization techniques \citep{Mazyavkina2021}. As a result, although neural-based methods are generally competitive for solving instances with up to 100 nodes, their generalization to larger instances remains limited. To address this, we incorporate an efficient local search algorithm after model inference, using a diverse set of search operators to iteratively refine solutions. This hybrid approach allows TuneNSearch to generalize more effectively than existing neural-based models across a range of VRP variants, in both randomly generated and benchmark instances. \textcolor{blue}{It achieves much lower relative performance deviations compared to other neural-based methods, while incurring only a small computational overhead.}

\section{Preliminaries}
\label{sec:prelim}

In this section we first describe the formulation of the MDVRP, followed by a brief overview of how neural-based methods formulate VRPs as a Markov decision process. After, we present other VRP variants featured in our work, along with their respective constraints.

\subsection{MDVRP description}
\label{sec:3.1}

The MDVRP is an extension of the classical VRP, in which multiple depots are considered. This problem can be stated as follows: a set $\mathcal{D}=\{d_1,d_2,\ldots,d_m\}$ of $m$ depots, and a set $\mathcal{C}=\{c_1,c_2,\ldots,c_n\}$ of $n$ customers are given. Combined, these sets form a set of nodes $\mathcal{V}=\mathcal{C} \cup \mathcal{D}$, where the total number of nodes is $n + m = g$. \textcolor{blue}{The edge set $\mathcal{E} = \left\{e_{ij} : i, j \in \mathcal{V}, \ i \neq j, \ (i, j) \notin \mathcal{D} \times \mathcal{D}\right\}$ represents the travel distances between distinct nodes in the problem, excluding direct connections between depots.} Each instance can be characterized by a graph $\mathcal{G}=\{\mathcal{V},\mathcal{E}\}$. A fleet of vehicles, each with a capacity $Q$, is dispatched from all depots to serve the customers. Each customer $c_i$ has a specific demand $\delta_i$ and must be visited exactly once by a single vehicle. Once a vehicle completes its route, it must return to its starting depot. \textcolor{blue}{The solution $\tau$ represents the sequence of nodes visited in the problem. It consists of multiple routes, where each route corresponds to the set of nodes visited by an individual vehicle.} In other words, $\tau$ captures the complete routing plan, breaking it down into distinct routes assigned to different vehicles. A solution is feasible as long as the capacity of each vehicle is not exceeded, and each customer is served exactly once. \textcolor{blue}{The objective is to find the optimal solution $\tau^\ast$ that minimizes the total distance traveled by all vehicles.}

\subsection{VRPs as a Markov decision process}
\label{sec:3.2}

Most existing neural-based models for VRPs use the reinforcement learning framework for training \citep{Sutton1998}. \textcolor{blue}{In our approach, solutions are built one route at a time. At the start of each route, the agent selects a depot from which the vehicle will depart. The agent then starts to append customers to that route, and the route remains active until the agent decides to close it by selecting the starting depot again. In this way, the agent manages each route independently, ensuring that all assignments for one route are completed before initiating another one. Once a route is closed, a new route is initialized from a depot (either the same or a different one) chosen by the agent, and this process continues until all customers have been assigned.} This approach can be formulated as a Markov decision process, which consists of the following key components:

\textbf{State:} The state is an observation received by the reinforcement learning agent which represents the current situation of the environment. \textcolor{blue}{Here, the environment is the system in which the agent operates, which in this case is a VRP instance. At each timestep $t$, the state $s_t$ includes embeddings of node features, which are vector representations capturing the relevant properties of each node in a continuous and numerical form, suitable to be processed by machine learning models. These embeddings are generated by an encoder, a neural network that transforms raw node features into an informative latent space. In addition to the node embeddings, $s_t$ also contains contextual information about the current partial solution, such as the vehicle’s remaining load, the total route length so far, and the embedding of the last visited node.}

\textbf{Action:} Upon receiving the state $s_t$, the agent selects the next action $a_t$ based on the current state. \textcolor{blue}{Actions can be of two types:
\begin{itemize}
	\item Depot node: chosen either to initialize a new route (by selecting the depot from which the vehicle departs) or to terminate the active route (by returning the vehicle to its starting depot).
	\item Customer node: chosen to extend the active route by visiting an unserved customer, subject to feasibility constraints. 
\end{itemize}
To enforce feasibility, a masking mechanism is applied at each timestep to dynamically restrict the action space. For customers, this implies that nodes already visited and those whose demand exceed the remaining vehicle capacity are masked out from the action selection process. For depot nodes, availability depends on the stage of the solution. Before initializing a new route, all depots are available, because the agent must decide to which depot the next route will correspond to. At this point, customers cannot be selected. Once a depot is chosen, the feasible customers become available, along with the chosen depot itself (to allow the agent to finish a route whenever it deems appropriate). When a route is closed, all depots become available again, enabling the agent to decide where the next route will begin. We note that in non-MDVRP instances, where only a single depot exists, an action corresponding to a depot merely indicates the start or end of a route, as the depot itself is always fixed.}

\textbf{State transition:} \textcolor{blue}{The state transition describes how the environment evolves as the agent picks actions. At each timestep $t$, the agent transitions from state $s_t$ to the next state $s_{t+1}$ based on the action $a_t$. If a customer is selected, it is appended to the current active route. Its demand is marked as fully served, the vehicle’s remaining capacity is updated, and the mask excludes that customer from future actions. If the action corresponds to the depot that initiated the active route, the route is closed and the vehicle returns to its departing depot. If, however, the depot is selected at the start of a new route, it sets the route’s origin. In both cases, the masking mechanism is updated following the previously explained procedure.}

\textbf{Reward:} \textcolor{blue}{The reward is a scalar value that tells the agent how good or bad its action was in a given state, acting as a feedback signal to guide the learning process towards optimizing a specific objective.} In the context of VRPs, at each timestep t, the agent receives a reward $r_t\ =\ -cost(s_{t+1},\ \ s_t)$ which reflects the negative distance traveled between states $s_t$ and $s_{t+1}$. \textcolor{blue}{Once the entire solution is completed, the cumulative reward $R\ =\ -cost(\tau)$ represents the negative total distance traveled in $\tau$.} The agent’s objective is to maximize its total cumulative reward, which aligns with minimizing the total distance traveled.

\textbf{Policy:} To maximize the total cumulative reward, the agent learns a policy, parametrized by an attention-based neural network (policy network) with parameters $\theta$. This policy is a function, or model, that maps states to actions. In essence, the agent learns a heuristic to determine how it should select the next action $a_t$ given the current state $s_t$, which is why neural-based methods are often referred to as neural heuristics. At each timestep $t$, the policy network takes as input the state $s_t$ and outputs the probabilities of visiting each node next. The agent then selects the node greedily (i.e., the node with the highest probability) or by sampling (choose an action stochastically based on the probabilities). \textcolor{blue}{This process continues until the full solution $\tau$ is constructed. The probability of constructing a solution $\tau$ can be expressed as $p_\theta(\tau|\mathcal{G})\ =\ \prod_{t=1}^{Z}p_\theta(a_t|\mathcal{G},\ a_{<t})$, where $a_t$ denotes the selected node, $a_{<t}$ the current partial solution and $Z$ is the maximum number of steps.} To train the model, most works use the REINFORCE algorithm \citep{Williams1992}, which is explained in more detail in Section~\ref{sec:4.2}.

Different from existing approaches, TuneNSearch initially pre-trains the policy network specifically on MDVRP data, which allows the model to establish a solid foundation of knowledge transferable to both single- and multi-depot variants. Then, for each VRP variant, the parameters from the pre-training phase are loaded into the model and a short fine-tuning phase is performed. In this phase, the model undergoes further training using data specific to each VRP variant. \textcolor{blue}{This fine-tuning phase is more efficient than training a new model from the beginning for each variant, as the model benefits from previously acquired transferable knowledge.}

\subsection{VRP variants}
\label{sec:3.3}

The VRP variants solved by TuneNSearch are: i) \textit{CVRP}: considers a single depot, in contrast to the multiple depots considered in MDVRP; \textcolor{blue}{ii) \textit{VRP with backhauls} (VRPB): in the classic CVRP, a vehicle departs the depot loaded, and each time it visits a customer, it delivers goods, thereby decreasing its load as the route progresses. These customers are known as linehaul customers. In the VRPB, some customers require picking up goods (backhaul customers), which increase the vehicle’s load. In this paper, we consider a mixed VRPB, where linehaul and backhaul customers can be visited in any order, allowing a vehicle to alternate between deliveries and pickups within the same route. However, at all times, the vehicle’s load must remain within its maximum capacity limit. This condition implies, for example, that if a vehicle leaves the depot fully loaded to its maximum capacity, it cannot serve a backhaul customer as its first stop, since this would immediately exceed its maximum capacity $Q$;}  iii) \textit{VRP with duration limit} (VRPL): in this variant, the length of each route cannot surpass a predefined threshold limit; iv) \textit{Open VRP} (OVRP): in the OVRP, vehicles do not need to return to the depot after completing their route;  v) \textit{VRP with time windows} (VRPTW): in the VRPTW, each node has a designated time window, during which service must be made, as well as a service time, representing the time needed to complete service at that location; vi) \textit{Traveling salesman problem} (TSP): the TSP is a simplified form of the CVRP that involves only a single route. In this problem, there is no depot, and nodes have no demands. Each node must be visited exactly once, and the vehicle (or salesman) must return to the starting node at the end of the route. We note that the constraints outlined above can also be applied to the MDVRP, instead of the CVRP. For more details on the generation of instances specific to all the described variants, we refer readers to \ref{sec:appB}.

\section{Methodology}
\label{sec:method}

In this section, we formally describe our proposed method, detailing each of the following components: i) Design of a novel model architecture combining POMO and the E-GAT encoder; ii) Implementation of a pre-training and fine-tuning framework designed to facilitate adaptation to the most common VRP variants; iii) Integration of a local search algorithm which applies different search operators to iteratively improve the solutions found by the neural-based model.

\subsection{Model architecture}
\label{sec:4.1}

\begin{figure*}[t]
	\begin{center}
		\includegraphics[width=\textwidth]{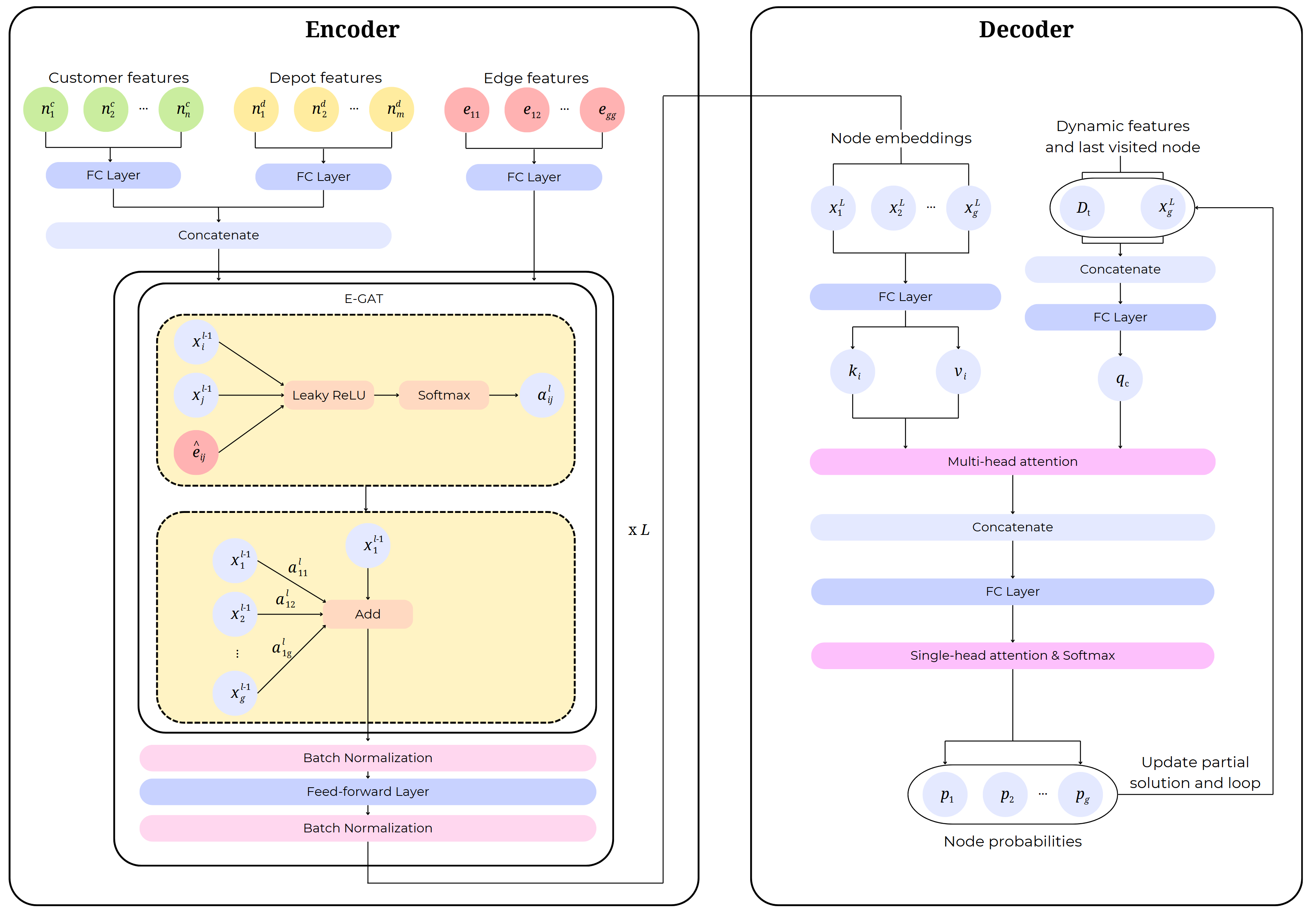}
		\caption{Encoder-decoder structure of TuneNSearch.} \label{fig1}
	\end{center}
\end{figure*}

Below we outline the architecture of TuneNSearch, which is built upon POMO. Our method incorporates the E-GAT in the encoder, which has previously shown performance improvements over the attention model \citep{Lei2022}. Unlike standard attention-based encoders that rely solely on node coordinates, the E-GAT extends the original graph attention network \citep{Veličković2018} by incorporating the information of edges $e_{ij}\in\ \mathcal{E} \ \forall \ i,\ j\ \in\left\{0,\ \ldots,\ g\right\}$. These enhancements allow the model to better capture the information of graph structures, deriving efficient representations and more accurate attention coefficients. When paired with POMO’s multiple starting nodes sampling strategy, this enhanced encoding allows the model to evaluate diverse solution trajectories from different starting points with more precision. The distance-aware attention mechanism ensures that each rollout receives more contextually relevant information, helping the model converge to higher-quality solutions. To the best of our knowledge, this is the first time the E-GAT encoder is combined with POMO. Fig.~\ref{fig1} presents an illustration of the encoder-decoder structure of TuneNSearch. The model first encodes the features of depots, customers, and edge distances using an E-GAT. \textcolor{blue}{The resulting embeddings, along with contextual information about the partial solution, are then passed through a multi-head attention (MHA) layer.} Finally, a single-head attention (SHA) layer, followed by a softmax function, calculates the probability of selecting each node next.

\subsubsection{Encoder details}
\label{sec:4.1.1}

Our encoder first embeds the features of all nodes in the problem to a $h_x$-dimensional vector space through a fully connected layer. Specifically, each depot $d_j \in \mathcal{D}$ is characterized by features $n_j^d$, which include its two-dimensional coordinates. On the other hand, each customer $c_i \in \mathcal{C}$ has features $n_i^c$, which not only contain its coordinates, but also the demand information and the early and late time windows. These features are passed through embedding layers separately for depots and customers. The result is a set of embedded vectors \{$\widehat{n}_j^d\ \in\ \mathbb{R}^{h_x}\ | j = 1, ..., m$\} for depots and \{$\widehat{n}_i^c\ \in\ \mathbb{R}^{h_x}\ | i = 1, ..., n$\} for customers, as shown in Equations~\ref{eq:1} and \ref{eq:2}. These vectors are then stacked to form $E^d \in \mathbb{R}^{m \times h_x}$ and $E^c \in \mathbb{R}^{n \times h_x}$. After, they are concatenated into $x^{\left(0\right)} \in \mathbb{R}^{g\ \times\ h_x}$, as demonstrated in Equation~\ref{eq:3}. \textcolor{blue}{Here, $x^{(0)}$ represents all nodes in the latent $h_x$-dimensional vector.} We also embed the edge features, which represent the Euclidean travel distances $e_{ij},\ i,\ j\ \in\ \left\{1,\ 2,\ \dots,\ g\right\}$, into a $h_e$-dimensional vector space, as described in Equation~\ref{eq:4}.
\begin{equation}
	\widehat{n}_i^c = A_0 n_i^c + b_0, \quad \forall i \in \{1, \dots, n\}
	\label{eq:1}
\end{equation}
\begin{equation}
	\widehat{n}_j^d\ =\ \left(A_1n_j^d\ +\ b_1\right),\ \forall j\ \in\ \left\{1,\ \ldots,\ m\right\}
	\label{eq:2}
\end{equation}
\begin{equation}
	x^{(0)} = concat(E^d, E^c) = \{x_1^{(0)}, \dots, x_g^{(0)}\}
	\label{eq:3}
\end{equation}
\begin{equation}
	\hat{e}_{ij} = \left(A_2 e_{ij} + b_2 \right), \quad \forall i, j \in \{1, 2, \dots, g\}
	\label{eq:4}
\end{equation}

After this initial transformation, both $x^{(0)}$ and $\hat{e} = {\hat{e}_{11},\hat{e}_{12},...,\hat{e}_{gg}}$ are used as inputs to an E-GAT module with $L$ layers. Here, the attention coefficient $\alpha_{ij}^l$ indicates the influence of the features of the node indexed by $j$ on the node indexed by $i$, at layer $l \in \left\{1,2,...,L\right\}$. The coefficient $\alpha_{ij}^l$ can be calculated according to Equation~\ref{eq:5}:
\begin{equation}
	\alpha_{ij}^l = \frac{exp(LeakyReLU(a^{l^T}[W_{1}^l(x_{i}^{(l-1)} || x_{j}^{(l-1)} || \hat{e}_{ij})])}{\sum\limits_{k=0}^{g} exp(LeakyReLU(a^{l^T}[W_{1}^l(x_{i}^{(l-1)} || x_{k}^{(l-1)} || \hat{e}_{ik})])}
	\label{eq:5}
\end{equation}
where $a^l$ and $W_1^l$ are learnable weight matrices.

Next, the node representations are updated using the weighted sum of neighboring nodes, followed by a residual connection and batch normalization:

\begin{equation}
	\tilde{x}_i^{(l)} = \text{BN}^l\left(x_i^{(l-1)} + \sum\limits_{k=0}^{g} \alpha_{ik}^l W_{2}^l x_{k}^{(l-1)} \right)
	\label{eq:6}
\end{equation}
where $W_2^l$ is another learnable matrix, and \text{BN} denotes batch normalization.

Finally, the output of this block is passed through a feed-forward layer, again followed by batch normalization:

\begin{equation}
	x_i^{(l)} = \text{BN}^l\left(\tilde{x}_i^{(l)} + \text{FF}^l(\tilde{x}_i^{(l)})\right)
	\label{eq:7}
\end{equation}

This process is repeated for all $L$ layers to obtain the final encoded node representations. \textcolor{blue}{In essence, the encoder generates node embeddings that capture the important features of each node, providing a compact summary of the entire problem instance. These embeddings are computed once by the encoder, before the decoder begins the solution construction process.}

\subsubsection{Decoder details}
\label{sec:4.1.2}

Following the approach proposed by \citet{Kool2019}, we apply an MHA layer followed by a SHA layer for the decoder. First, the decoder takes the initial node embeddings $x_i^{(L)}$ (we omit the $(L)$ term for better readability) and sets the keys and values for all $H$ heads of the MHA, as indicated in Equations~\ref{eq:8} and ~\ref{eq:9}:
\begin{equation}
	v_i = W^Vx_i, \forall i \in \left\{1, 2, ..., g\right\}
	\label{eq:8}
\end{equation}
\begin{equation}
	k_i = W^Kx_i, \forall i \in \left\{1, 2, ..., g\right\}
	\label{eq:9}
\end{equation}
where $W^V, W^K \in \mathbb{R}^{h_v \times h_x}$ are learnable matrices and $h_v = (h_x / H)$, with $v_i, k_i \in \mathbb{R}^{h_v}$.

To generate the query vector $q_c$, the embeddings of the currently selected node ($x_{i_t}$, at timestep $t$) are concatenated with dynamic features $D_t$, as described in Equation~\ref{eq:10}.
\begin{equation}
	q_c = W^Q concat(x_{i_t}, D_t)
	\label{eq:10}
\end{equation}
where $W^Q \in \mathbb{R}^{h_v \times h_x}$ is a learnable matrix. The features $D_t$ include the vehicle load at timestep $t$, the elapsed time, the length of the current route and a Boolean to indicate whether routes are open or not. For the TSP, we do not perform this concatenation operation, as the problem is solely defined by node coordinates. Therefore, we exclude the 4 neurons associated with the dynamic features $D_t$.

The node compatibilities $u_{ci}, \forall i \in \left\{1,2,…,g\right\}$ are then calculated through the query vector $q_c$ and the key vector $k_i$, as shown in Equation~\ref{eq:11}:

\begin{equation}
	u_{ci} = C.tanh(q_c^Tk_i)
	\label{eq:11}
\end{equation}
where the results are clipped within $[-C,C]$. Furthermore, the compatibility of infeasible nodes is set to $-\infty$ to guarantee that only feasible solutions are generated. Lastly, the probability $p_i, \forall i \in \left\{1,2,…,g\right\}$ of each node is computed through a softmax function, as indicated in Equation~\ref{eq:12}.

\begin{equation}
	p_i = \frac{e^{u_{ci}}}{\sum\limits_{j=1}^{g} e^{u_{cj}}}
	\label{eq:12}
\end{equation}

\textcolor{blue}{This process continues indefinitely, with the decoder selecting one node at a time, until a complete solution is constructed for each instance.}

\subsection{Model training}
\label{sec:4.2}

To train our model, we used the REINFORCE algorithm \citep{Williams1992}, which is a fundamental policy gradient method used in reinforcement learning. In particular, we use the REINFORCE with shared baselines algorithm, following the approach of POMO \citep{Kwon2020}, where multiple solutions/trajectories are sampled with $N$ different starting nodes. \textcolor{blue}{A trajectory refers to the full sequence of actions or decisions that the model takes to construct a complete solution. In the context of VRPs, a trajectory can essentially be interpreted as a solution. This means that, in a batch with $B$ different instances, the model simultaneously constructs $N$ trajectories per instance, generating $B \times N$ solutions in parallel. For each trajectory, the first action is deterministically chosen, after which the decoder will continue to build each trajectory as explained before, according to the calculated node probabilities. The rationale behind this design is to encourage diversity in the generated solutions, since each customer has the opportunity to serve as a starting point. As a result, the model explores a broader range of possible solutions, improving its coverage of the solution space.}

Once all $B \times N$ trajectories have been constructed, we compute the total rewards $R(\tau^1,…,\tau^N )$ for each solution, and use gradient ascent to maximize the total expected return $J$, as indicated in Equation~\ref{eq:13}:

\begin{equation}
	\nabla_{\theta}J(\theta) = \frac{1}{BN} \sum\limits_{i=1}^{B} \sum\limits_{j=1}^{N} (R(\tau_i^j) - b_i) \nabla_{\theta} \log p_{\theta} (\tau_i^j)
	\label{eq:13}
\end{equation}
where $\theta$ is the set of model parameters and $p_{\theta} (\tau_i^j)$ is the probability of trajectory $\tau_i^j$ being selected. Furthermore, $b_i$ is a shared baseline calculated according to Equation~\ref{eq:14}:

\begin{equation}
	b_i = \frac{1}{N} \sum\limits_{j=1}^{N} R(\tau_i^j), \forall i \in \left\{1, ..., B\right\}
	\label{eq:14}
\end{equation}

Algorithm~\ref{algo:1} outlines the REINFORCE with shared baselines algorithm in more detail. First, the policy network is initialized with a random set of parameters $\theta$. Then, training begins by looping over $E$ epochs, each comprising $T$ steps. At each step, a batch of $B$ random training instances is sampled, and for each instance, $N$ starting nodes are selected. For all experiments, we set $N$ as $g - 1$, consistent with prior work \citep{Kwon2020,Li2024}. From these nodes, trajectories are sampled through the model architecture explained in Section~\ref{sec:4.1}. \textcolor{blue}{In our implementation, node indices are assigned such that depots occupy the first positions, followed by customers (e.g., 0, 1 and 2 for a problem with three depots). Like prior work \citep{Li2024}, depot 0 is treated as the implicit starting point of the first route unless the deterministic starting node itself corresponds to another depot (e.g., index 1 or 2). For example, in a problem with 3 depots and 50 customers, we set $N = 52$: two trajectories start from depots with indices 1 and 2, while the remaining 50 trajectories start from customers with indices 3–52, all initially departing from depot 0. Although one could consider all depot–customer pairs as starting states (yielding 150 trajectories in this case), the computational cost would be prohibitive. Importantly, initializing multiple trajectories at depot 0 does not bias the solutions towards a single depot. The reason is that depot assignment is not fixed by the initialization: throughout decoding, the model explicitly decides when to close a route and from which depot the next vehicle should depart. Thus, even if the first route in a trajectory begins at depot 0, subsequent routes may start at other depots whenever the policy deems it as beneficial.} 

After this, the shared baseline is calculated (Equation~\ref{eq:14}), which is used to compute the policy gradients  $\nabla_\theta J(\theta)$ (Equation~\ref{eq:13}). Lastly, the set of parameters $\theta$ is updated through gradient ascent, scaled by a learning rate $\eta$.

\begin{algorithm}
	\caption{REINFORCE with shared baselines \citep{Kwon2020}}
	\label{algo:1}
	\begin{algorithmic}[1]
		\REQUIRE Number of epochs $E$, batch size $B$, steps per epoch $T$
		\STATE Initialize policy network with parameters $\theta$
		\FOR{$epoch = 1$ \TO $E$}
		\FOR{$step = 1$ \TO $T$}
		\STATE Randomly sample set $\mathcal{S}=\{\mathcal{G}_1,\mathcal{G}_2,...,\mathcal{G}_B\}$ with $B$ training instances
		\STATE Select $N$ starting nodes for each instance $\mathcal{G}_i$, $i \in \{1, ..., B\}$
		\STATE Using the selected starting nodes, sample trajectories $\tau_i^j$, $\forall i \in \{1, ..., B\}, \forall j \in \{1, ..., N\}$
		\STATE $b_i \gets \frac{1}{N} \sum_{j=1}^{N} R(\tau_i^j)$, $\forall i \in \{1, ..., B\}$
		\STATE $\nabla_{\theta}J(\theta) \gets \frac{1}{BN} \sum_{i=1}^{B} \sum_{j=1}^{N} (R(\tau_i^j) - b_i) \nabla_{\theta} \log p_{\theta} (\tau_i^j)$
		\STATE $\theta \gets \theta + \eta \nabla_\theta J(\theta)$
		\ENDFOR
		\ENDFOR
	\end{algorithmic}
\end{algorithm}

\subsection{Pre-training and fine-tuning process}

With the architecture described above, we pre-train a backbone model for 100 epochs, similarly to the framework proposed by \citet{Lin2024}. However, rather than training on TSP data, we use MDVRP instances. This choice is motivated by the higher complexity of the MDVRP, which enables the model to learn richer node representations in comparison to the TSP.  \textcolor{blue}{Pre-training on MDVRP data also offers advantages over using CVRP data, since the model is exposed to multi-depot structures.}

Additionally, in contrast to \citet{Lin2024}, during pre-training we incorporate all dynamic features $D_t$ relevant across all VRP variants considered. Rather than introducing problem-specific modules, this approach allows us to fine-tune the model for each VRP variant without modifying the neural network architecture. This avoids the loss of potentially useful knowledge when transitioning between variants. The only structural change occurs when adapting the model for the TSP, as it is the simplest routing problem we address. \textcolor{blue}{The TSP differs from other VRP variants since it involves only node coordinates, with no demands, time windows, or additional constraints. In fact, the only necessary features to encode a TSP instance are the 2D coordinates of each node in the problem and the edge distances between each pair of nodes. Thus, for TSP-specific fine-tuning, we make the following adjustments:
\begin{enumerate}
	\item We begin by initializing the model with the parameters obtained from the MDVRP pre-training phase, leaving the original deep neural network architecture intact;
	\item Next, we discard the layers of the model associated with customer-specific features, since demands and time windows are not part of the TSP. Only the layers responsible for encoding node coordinates are retained, which correspond to the depot-embedding layers in the MDVRP model. We also keep the layers responsible for encoding the edge features, as spatial relationships between nodes remain essential in the TSP. For clarity, when we say that these layers are discarded, we mean that they are removed from the model architecture rather than being merely ignored;
	\item We also modify the decoder to remove the dynamic features tied to VRP constraints (vehicle load, elapsed time, route length and the open-route indicator);
	\item Finally, the solution decoding process follows TSP logic. At each step, the agent selects the next node to be visited by the salesman, and the first visited node is revisited only at the end of the trajectory. The masking mechanism ensures that nodes already visited cannot be selected again.
	\end{enumerate}}

\begin{figure*}[ht]
	\begin{center}
		\includegraphics[width=\textwidth]{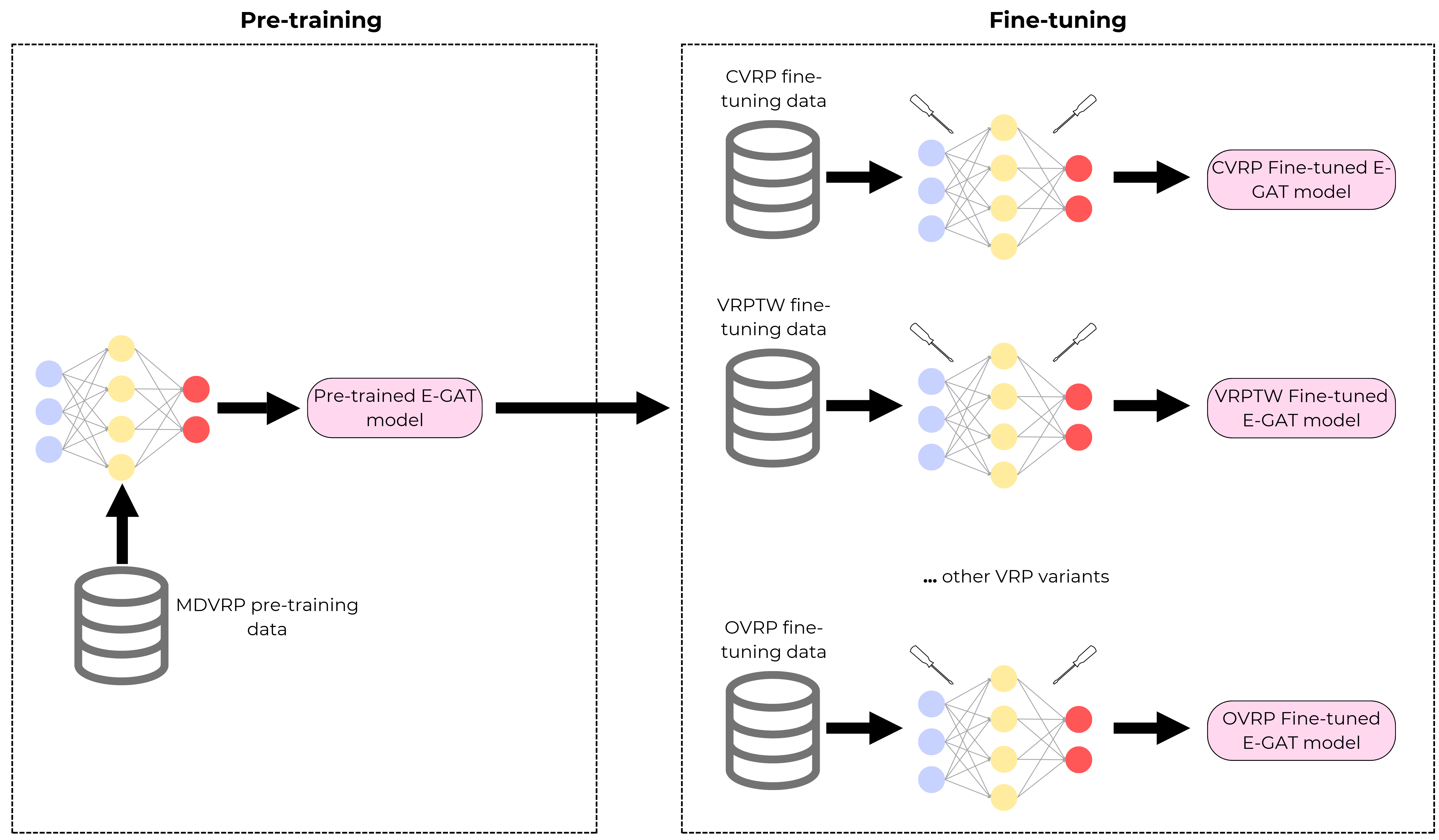}
		\caption{TuneNSearch pre-training and fine-tuning overview.} \label{fig2}
	\end{center}
\end{figure*}

For the fine-tuning phase, we load the parameters from the pre-trained MDVRP model and train it for an additional 20 epochs on randomly generated instances, of each specific VRP variant. \textcolor{blue}{We use the same hyper-parameters as in the initial training phase.} The dynamics of the pre-training and fine-tuning stages of TuneNSearch are illustrated in Fig.~\ref{fig2}. We note that besides the main variants described in Section~\ref{sec:3.3}, TuneNSearch can be extended to any combination of VRP constraints. For example, it can handle combinations like the MDVRP with time windows, open routes and backhauls, allowing it to tackle more complex routing problems.

\subsection{Local search algorithm}

Despite the recent interest in using neural-based techniques to solve VRPs, there remains a limited integration between operations research and machine learning methods in the current literature. Bridging this gap offers an opportunity to combine the strengths of both fields, enabling the development of more powerful and efficient algorithms. To refine the solutions obtained by the neural-based model, we designed an efficient local search algorithm employed after inference, inspired by different existing methods. \textcolor{blue}{In this section, we describe each component of our local search.}

\textbf{Search operators and granular neighborhood:} The core of our local search algorithm lies in applying a set of different operators within restricted neighborhoods of predefined size. Following \citet{Vidal2022}, we define the neighborhood size as 20, which allows for an efficient exploration of the granular neighborhood. A granular neighborhood is a restricted subset of the solution space that is defined based on proximity criteria. Rather than evaluating moves across the entire problem domain, the search is confined to carefully selected neighborhoods of limited size, where meaningful improvements are more likely to be found. Moves are restricted to node pairs $(a,b)$, where $b$ is one of the 20 closest nodes to $a$. Each move is evaluated within different neighborhoods, and any improvement in the cost function is immediately applied. The search procedure terminates when all applicable operators and moves have been applied without yielding any further improvement in the cost function. We adopt the same set of operators as \citet{Wouda2024}, which were selected due to their ability to effectively explore the solution space while maintaining computational efficiency. \textcolor{blue}{ The first three operators focus on node-level modifications, attempting moves between pairs of nodes whenever one node lies within the restricted neighborhood of the other. In contrast, the last two operators function at the route level, exchanging or repositioning customers between routes irrespective of their granular neighborhood.} Each operator targets different aspects of the solution structure, ensuring a well-balanced and diverse set of moves to improve the solution quality. The chosen operators include:

\begin{itemize}
	\item \textbf{(X, M)-exchange:} involves swapping $X$ customers from a route (starting at a designated node), with a segment of $M$ customers (where $0 \leq M \leq X$) starting at a different node. Importantly, the two segments must not overlap, ensuring that the exchange modifies the segments without duplicating any customers.
	
	\item \textbf{MoveTwoClientsReversed:} involves selecting two customers from a given route and moving them to a different position in a reversed order, essentially functioning as a reversed (2, 0)-exchange.
	
	\item \textbf{2-OPT:} iteratively removes two edges from a route and reconnects the resulting paths in a different configuration.
	
	\item \textbf{RELOCATE*:} \textcolor{blue}{identifies and executes the best possible relocation of a single customer between two routes. It removes the customer from its current route and inserts it into the best position in the other route, such that the total cost is minimized (with respect to all other possible relocation moves).}
	
	\item \textbf{SWAP*:} \textcolor{blue}{evaluates and executes the most beneficial exchange of two customers between two routes, positioning each customer in the best possible location within the other route. Unlike a traditional swap, this operator exchanges two customers from different routes by inserting them into any position on the opposite route, rather than directly swapping them in place.}
\end{itemize}

\textbf{Local search algorithm workflow:} We now describe the overall workflow of the proposed local search algorithm (see Algorithm~\ref{algo:2}). It begins with an initial solution $\tau$, which is the best solution obtained from the E-GAT model. The algorithm assumes a fixed number of iterations $I$, and solutions are represented as sequences of visited nodes. The best distance ($BD$) is initialized as the cost of $\tau$, and the best solution ($BS$) is set to $\tau$. \textcolor{blue}{Running an iterated local search starting from a randomly initialized solution can be very computationally expensive, and may require many iterations to converge. By seeding the search with a high-quality solution generated by the E-GAT model, we can significantly reduce the search space, and thus concentrate the search on more promising regions, accelerating convergence (see Section~\ref{sec:5.4} for a comparison with randomly and greedily initialized solutions).} 

Before the iterative process starts, the algorithm explores the neighborhood of $\tau$ using the $Search$ function. This function applies all the operators and moves explained above to identify potential improvements to the solution. \textcolor{blue}{A more detailed explanation of this function is given below, in Algorithm~\ref{algo:3}.} If a better solution is found during this step, both $BD$ and $BS$ are updated accordingly. After that, in each iteration, to escape local minimum, an offspring solution ($OS$) is generated using the $Crossover$ function. This function takes as parents the current $BS$ and a randomly generated solution, created with the $MakeRandom$ function. Here, we use the selective route exchange crossover (SREX) operator proposed by \citet{Nagata2010}. Then, the neighborhood of $OS$ is explored through the $Search$ function. If the resulting solution is infeasible, the algorithm repairs it using the $Fix$ function, which applies the search operators again while considering only feasible moves. Finally, if the new solution’s cost is better than the current $BD$, the algorithm updates $BD$ and $BS$. This process is repeated for $I$ iterations.

\textcolor{blue}{The rationale behind the choice of always iterating the local search over $BS$ stems from balancing diversification and intensification. On the diversification side, the crossover operator combines the current best solution with a randomly generated solution, introducing more diversity and enabling the method to escape local optima. On the intensification side, anchoring the search around the best solution ensures that exploration remains focused on high-quality regions of the solution space, promoting steady progress toward better solutions. Iterating only from $BS$ risks getting trapped in local minimum, while iterating solely from random solutions would often lead the search into poor-quality regions. By combining both, we allow the algorithm to explore new structural patterns while preserving the advantages of high-quality routes already identified.}

\begin{algorithm}[ht]
	\caption{Local Search Algorithm}
	\label{algo:2}
	\begin{algorithmic}[1]
		\REQUIRE Number of iterations $I$, initial solution $\tau$
		\STATE $BD \gets Cost(\tau)$
		\STATE $BS \gets \tau$
		\STATE $\tau' \gets Search(\tau)$
		\IF{$Cost(\tau') < BD$}
		\STATE $BD \gets Cost(\tau')$
		\STATE $BS \gets \tau'$
		\ENDIF
		\FOR{$iteration = 1$ \TO $I$}
		\STATE $OS \gets Crossover(BS, MakeRandom())$
		\STATE $\tau \gets Search(OS)$
		\IF{\textbf{not} $IsFeasible(\tau)$}
		\STATE $\tau \gets Fix(\tau)$
		\ENDIF
		\IF{$Cost(\tau) < BD$}
		\STATE $BD \gets Cost(\tau)$
		\STATE $BS \gets \tau$
		\ENDIF
		\ENDFOR
		\RETURN $BS$
	\end{algorithmic}
\end{algorithm}

\textcolor{blue}{\textbf{Search function:} Algorithm~\ref{algo:2} provides a high-level view of the entire local search procedure. A key component of this procedure is the $Search$ function, which explores different neighborhoods and applies the operators described earlier. Following the design of \citet{Vidal2022}, the implementation of $Search$ is detailed in Algorithm~\ref{algo:3}. The function takes a solution $\tau$ as input and iteratively improves it into a solution $\tau'$. The search focuses first on node-level operators, i.e., (X, M)-exchange, MoveTwoClientsReversed and 2-OPT, which are applied in a randomized order at each function call to introduce stochasticity. For every node $a$ in the solution, the algorithm iterates over each node $b$ in its granular neighborhood, and evaluates each of the three operators on all admissible node pairs $(a, b)$. Whenever a move yields an improvement in the cost function, it is applied immediately and the solution is updated accordingly. This process repeats until a complete pass over all node pairs and operators produces no further improvements, at which point the node-level search terminates. Next, the function applies route-level operators (RELOCATE* and SWAP*), again in a different random order at each function call. It iterates over all route pairs in the solution and evaluates both operators. As before, any improving move is applied immediately and the solution is updated. Like the node-level search, this phase ends once all route pairs have been evaluated without improvement, completing the overall search process. Overall, the $Search$ function terminates only when no further locally improving move can be found.}

\begin{algorithm}[ht]
	\caption{Search Function}
	\label{algo:3}
	\color{black}
	\begin{algorithmic}[1]
		\REQUIRE Solution $\tau$
		\ENSURE Improved solution $\tau'$
		\STATE Initialize $\tau' \gets \tau$
		\STATE Randomly shuffle the order of node and route operators
		\STATE \textbf{// Node-level search}
		\REPEAT
		\STATE improvement $\gets$ false
		\FOR{each customer node $a$ in $\tau'$}
		\FOR{each node $b$ in the granular neighborhood of $a$}
		\FOR{each node operator $nop$}
		\IF{$nop(a,b)$ improves cost}
		\STATE Apply $nop(a,b)$ to $\tau'$ and update solution
		\STATE improvement $\gets$ true
		\ENDIF
		\ENDFOR
		\ENDFOR
		\ENDFOR
		\UNTIL {improvement = false}
		
		\STATE \textbf{// Route-level search}
		\REPEAT
		\STATE improvement $\gets$ false
		\FOR{each pair of routes $(\mathcal{R}_i, \mathcal{R}_j)$ in $\tau'$}
		\FOR{each route operator $rop$}
		\IF{$rop(\mathcal{R}_i, \mathcal{R}_j)$ improves cost}
		\STATE Apply $rop(\mathcal{R}_i, \mathcal{R}_j)$ to $\tau'$
		\STATE Update solution and update solution
		\STATE improvement $\gets$ true
		\ENDIF
		\ENDFOR
		\ENDFOR
		\UNTIL{improvement = false}
		
		\RETURN $\tau'$
	\end{algorithmic}
\end{algorithm}

\textcolor{blue}{\textbf{Fix function:} Within the local search algorithm workflow, the $Search$ function may occasionally produce infeasible solutions. This happens because the cost of a solution is evaluated using penalty coefficients for constraint violations. When constraints like vehicle capacity, time windows or route duration limits are violated, they are accounted for in the resulting cost function. However, in the case of the $Search$ function, we use very small penalty coefficients. As a result, infeasible solutions may be, on many occasions, more favorable than feasible ones, since the penalties are often outweighed  by the reduction in the total distance traveled. Allowing these temporary infeasibilities enables the search process to explore beyond strictly feasible regions of the solution space, reaching areas that would otherwise remain inaccessible.}

\textcolor{blue}{The $Fix$ function was introduced in our algorithm to repair these occasionally generated infeasible solutions. It does so by applying all previously described search operators to the infeasible solution, but this time with very large penalty coefficients (similar to prior studies, such as \citet{Vidal2022}). By imposing overly high penalties for constraint violations, the $Fix$ function makes the cost of infeasible solutions highly uncompetitive, ensuring that only feasible moves are applied. Therefore, all results reported in this paper are guaranteed to be feasible. Furthermore, even after applying the $Fix$ function, additional feasibility checks are performed over the solution it produces to ensure that only feasible solutions can be returned by our local search algorithm.}

\textcolor{blue}{\textbf{Crossover:} The crossover operator is central to our search, as it must balance two competitive objectives: exploring new regions of the search space, while also preserving promising solution structures. To achieve this balance, we use the SREX operator, whose design explicitly support both goals. This operator works as follows:
	\begin{enumerate}
		\item Two parents, $P_A$ and $P_B$, are given (in our case, the current best solution $BS$ and a randomly generated solution).
		\item Two offspring solutions, $OS_1$ and $OS_2$, are initialized as copies of $P_A$.
		\item Two sets of routes, $S_A$ (from $P_A$) and $S_B$ (from $P_B$), are randomly chosen.
		\item The routes present in $S_A$ are removed from $OS_1$. Customers that appear in $S_B$ but not in $S_A$ are also removed from $OS_1$.
		\item The routes present in $S_A$ are removed $OS_2$. The routes from $S_B$ are inserted into $OS_1$.
		\item The routes from $S_B$ are inserted into $OS_2$, while excluding from $S_B$ any customers that are in $S_B$ but not in $S_A$.
		\item Any remaining unserved customers are reinserted into each offspring based on the least costly insertions.
		\item Evaluate both offsprings created and return the one with the better objective value.
	\end{enumerate}
}

\textcolor{blue}{For comparison, the order crossover (OX), used for example in many well-known studies \citep{PRINS2004,VIDAL2013,ILHAN2021}, creates an offspring by copying a single segment of nodes from one parent and filling the remaining positions in the order they appear in the other parent. Although conceptually simple and straightforward, OX often disrupts effective routes, resulting in offspring of limited quality. In contrast, SREX generates offsprings by inheriting multiple different routes that are kept mostly intact. This ensures that high-quality routes are more likely to be retained in the offspring, rather than being discarded or disrupted. As a result, SREX produces offsprings that remain close to favorable areas of the search space, while still introducing enough diversity to help the algorithm escape local minimum.}

\textcolor{blue}{Beyond this theoretical advantage, SREX has also demonstrated strong empirical performance. Originally developed for the pickup and delivery problem with time windows, a task that closely resembles the VRP with backhauls and time windows, it demonstrated impressive performance compared to other well-known heuristics \citep{Bent2006,Ropke2006}, improving 146 out of 298 benchmark problems. Its design also makes it highly flexible, allowing it to be applied across a wide range of VRP variants.}

\textcolor{blue}{To further validate this choice, during the design phase of TuneNSearch, we conducted the experiments reported in Section~\ref{sec:5.1}, comparing the performance of our local search algorithm when using SREX versus OX. SREX consistently outperformed OX across all VRP variants, reinforcing its suitability in our work. For the aforementioned reasons, we rely on SREX in our algorithm, using OX solely for the TSP, as it is a problem in which solutions consist of a single sequence of nodes rather than distinct routes, making SREX inapplicable.}

\begin{figure*}[ht]
	\begin{center}
		\includegraphics[width=\textwidth]{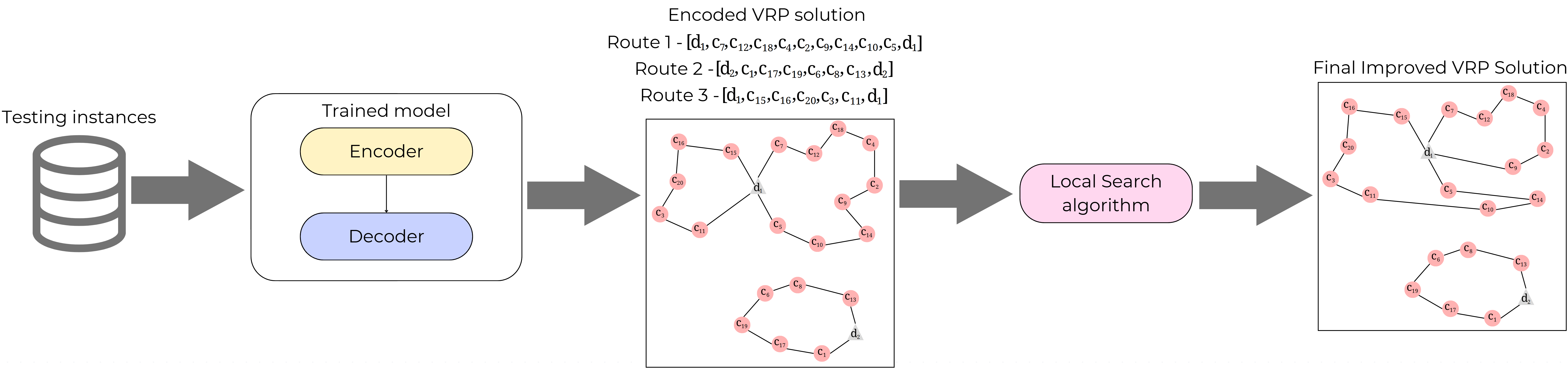}
		\caption{Overview of the inference process of TuneNSearch.} \label{fig3}
	\end{center}
\end{figure*}

Fig.~\ref{fig3} depicts an overview of the complete inference process of TuneNSearch, coupled with the local search algorithm.

\section{Experimental results}
\label{sec:experiments}

In this section, we verify and demonstrate the performance of TuneNSearch through a series of extensive experiments. In addition to the MDVRP, we consider the primary VRP variants described in Section~\ref{sec:3.3}. Our models were implemented using PyTorch \citep{Paszke2017}, and all neural-based experiments were conducted on a machine with 45GB of RAM, an Intel Xeon Gold 5315Y and an Nvidia Quadro RTX A6000. The baselines PyVRP and OR-Tools Routing library were executed on a machine with 32 GB of RAM and an Intel Core i9-13900. We note that the use of different computers does not affect the fairness of the comparison, as PyVRP and OR-Tools are not neural-based frameworks and therefore do not benefit from GPU acceleration.

\textbf{Baselines:} To evaluate the performance of TuneNSearch, we compared it against four state-of-the-art baseline approaches: PyVRP, an extension of the state-of-the-art hybrid genetic search (HGS) \citep{Vidal2022}, OR-Tools Routing library, MD-MTA and POMO. For PyVRP and OR-Tools, we set time limits of 300, 600 and 1800 seconds for problems of sizes $n = 20,50,100$, respectively. Both methods were parallelized across 32 CPU cores, following prior research \citep{Kool2019,Zhou2024}.

\textbf{Training and hyper-parameters:} We consider three MDVRP models trained on instances of sizes $n = 20,50,100$, where the number of depots ($m$) was set to 2, 3, and 4, respectively. Each model was trained for 100 epochs, with one epoch consisting of 320000 training instances for $n \in {20,50}$, and 160000 for $n = 100$. \textcolor{blue}{The batch size $B$ was chosen based on GPU memory limits to make training more efficient and faster. For smaller problems (with $n = 20$), we used $B = 2000$, since the instances were simpler and allowed a more effective GPU utilization. For larger problems, with 50 and 100 nodes, we reduced  $B$ to 500 and 120, respectively, since these instances required much more memory than the 20-node ones.} When fine-tuning our pre-trained MDVRP model, we consider 20 additional epochs for each VRP variant. For all baselines, we maintained the original hyper-parameters from their respective works. For TuneNSearch, we set the hidden dimension at $h_x  = 256$ and the hidden edge dimension at $h_e  = 32$. The number of encoder layers is $L = 5$, and the feed-forward layer in the encoder has a hidden dimension of 512. The number of heads in the MHA decoder is $H = 16$, and the tanh clip size was set to 10, according to \citet{Bello2017}. We employed the Adam optimizer \citep{Kingma2015} with a learning rate of $\eta =  10^{-4}$. \textcolor{blue}{Across all experiments, we set the number of starting nodes $N$ to $g - 1$. Essentially, this means that for every instance solved, $g - 1$ different solutions are generated following the procedure explained in section~\ref{sec:4.2}. This choice follows a convention in the literature, from which most studies adopt the same value \citep{Kwon2020,Lin2024,Li2024}. It allows a broad exploration of the solution space, while keeping the total training time manageable.} Finally, a sensitivity analysis of key hyper-parameters is provided in \ref{sec:appA}.

\subsection{Computational results on randomly generated instances}
\label{sec:5.1}

Table~\ref{table1} provides a comparison between our pre-trained MDVRP model and MD-MTA, as well as between POMO (specialized for each specific VRP variant) and our fine-tuned models. The evaluation is based on datasets of 1280 randomly generated instances for each variant and instance size pair, and considers three key metrics: the average total distance traveled (Obj.), the average relative percentage deviation (RPD) compared to the best solutions found across all evaluated methods, and the computational time (Time) taken to solve all instances. \textcolor{blue}{The RPD for each method is computed as the average, over all instances in a dataset $D$, of the percentage deviation from the best-performing method on each instance (i.e., the method achieving the lowest objective on that instance):
\begin{equation}
	 RPD = \frac{1}{|D|}\sum_{i=1}^{|D|}\frac{z_i - z_{best, i}}{z_{best, i}} \times 100\%
	 \label{eq:15}
\end{equation}
where $z_i$ is the objective value of the evaluated method on instance $i$, and $z_{best, i}$ is the smallest objective value among all evaluated methods on instance $i$. The best results, indicated in \textbf{bold}, correspond to the lowest Obj. values and the lowest RPD (relative to HGS-PyVRP).}

For all neural-based methods, we used a greedy decoding with x8 instance augmentation \citep{Kwon2020}, except for MD-MTA, which we applied the enhanced depot rotation augmentation designed by \citet{Li2024}. Additionally, we evaluated TuneNSearch under two other configurations: i) with the local search applied post-inference (labeled \textit{ls.}); ii) with local search but without x8 instance augmentation (labeled \textit{ls. + no aug.}). In both configurations, the local search was performed for 50 iterations.

\begin{table*}[!ht]
	\centering
	\caption{Experimental results on all VRP variants (* represents 0.000 \% RPD).} 
	\label{table1}
	\small
	\resizebox{\textwidth}{!}{%
		\begin{tabular}{cc|ccc|ccc|ccc}
			\hline
			& \multirow{2}{*}{Method} & \multicolumn{3}{c|}{$n$ = 20} & \multicolumn{3}{c|}{$n$ = 50} & \multicolumn{3}{c}{$n$ = 100} \\
			&  & Obj. & RPD & Time (m) & Obj. & RPD & Time (m) & Obj. & RPD & Time (m) \\
			\hline
			\multirow{6}{*}{MDVRP} & HGS-PyVRP & 4.494 & * & 200.134 & 7.489 & * & 401.243 & 11.345 & * & 1200.436 \\
			& OR-Tools & \textbf{4.494} & \textbf{*} & 200.026 & 7.523 & 0.454 \% & 400.043 & 11.637 & 2.574 \% & 1200.160 \\
			& MD-MTA & 4.548 & 1.202 \% & 0.104 & 7.679 & 2.537 \% & 0.237 & 11.797 & 3.984 \% & 0.944 \\
			& Ours & 4.518 & 0.534 \% & 0.055 & 7.637 & 1.976 \% & 0.146 & 11.744 & 3.517 \% & 0.535 \\
			& Ours (ls.) & 4.495 & 0.022 \% & 0.263 & \textbf{7.521} & \textbf{0.427 \%} & 1.017 & \textbf{11.521} & \textbf{1.551 \%} & 3.270 \\
			& Ours (ls. + no aug.) & 4.496 & 0.045 \% & 0.241 & 7.535 & 0.614 \% & 0.908 & 11.582 & 2.089 \% & 2.784 \\
			\hline
			\multirow{6}{*}{CVRP} & HGS-PyVRP & 4.977 & * & 200.128 & 9.392 & * & 401.465 & 16.124 & * & 1200.323 \\
			& OR-Tools & 4.977 & * & 200.023 & 9.468 & 0.809 \% & 400.036 & 16.613 & 3.033 \% & 1200.038 \\
			& POMO & 4.991 & 0.281 \% & 0.040 & 9.502 & 1.171 \% & 0.089 & 16.531 & 2.524 \% & 0.296 \\
			& Ours & 4.992 & 0.301 \% & 0.050 & 9.494 & 1.086 \% & 0.133 & 16.430 & 1.898 \% & 0.484 \\
			& Ours (ls.) & 4.977 & * & 0.257 & \textbf{9.422} & \textbf{0.319 \%} & 1.045 & \textbf{16.303} & \textbf{1.110 \%} & 3.310 \\
			& Ours (ls. + no aug.) & 4.978 & 0.020 \% & 0.244 & 9.435 & 0.458 \% & 0.964 & 16.370 & 1.526 \% & 2.920 \\
			\hline
			\multirow{6}{*}{VRPB} & HGS-PyVRP & 4.551 & * & 200.119 & 8.119 & * & 401.439 & 13.450 & * & 1200.328 \\
			& OR-Tools & \textbf{4.551} & \textbf{*} & 200.023 & \textbf{8.144} & \textbf{0.308 \%} & 400.031 & 13.766 & 2.349 \% & 1200.128 \\
			& POMO & 4.583 & 0.703 \% & 0.041 & 8.274 & 1.909 \% & 0.085 & 13.938 & 3.628 \% & 0.294 \\
			& Ours & 4.578 & 0.593 \% & 0.043 & 8.285 & 2.045 \% & 0.128 & 13.867 & 3.100 \% & 0.451 \\
			& Ours (ls.) & 4.554 & 0.066 \% & 0.243 & 8.159 & 0.493 \% & 0.996 & \textbf{13.660} & \textbf{1.561 \%} & 3.481 \\
			& Ours (ls. + no aug.) & 4.554 & 0.066 \% & 0.212 & 8.168 & 0.604 \% & 0.915 & 13.717 & 1.985 \% & 2.861 \\
			\hline
			\multirow{6}{*}{VRPL} & HGS-PyVRP & 4.998 & * & 200.124 & 9.349 & * & 401.492 & 16.138 & * & 1200.318 \\
			& OR-Tools & 4.998 & * & 200.024 & 9.420 & 0.759 \% & 400.037 & 16.644 & 3.135 \% & 1200.045 \\
			& POMO & 5.022 & 0.480 \% & 0.084 & 9.457 & 1.155 \% & 0.134 & 16.507 & 2.287 \% & 0.364 \\
			& Ours & 5.017 & 0.380 \% & 0.099 & 9.451 & 1.091 \% & 0.181 & 16.448 & 1.921 \% & 0.556 \\
			& Ours (ls.) & 4.998 & * & 0.295 & \textbf{9.378} & \textbf{0.310 \%} & 1.084 & \textbf{16.324} & \textbf{1.153 \%} & 3.366 \\
			& Ours (ls. + no aug.) & 4.999 & 0.020 \% & 0.276 & 9.391 & 0.449 \% & 0.976 & 16.394 & 1.586 \% & 2.976 \\
			\hline
			\multirow{6}{*}{OVRP} & HGS-PyVRP & 3.480 & * & 200.138 & 6.150 & * & 401.387 & 9.949 & * & 1200.345 \\
			& OR-Tools & \textbf{3.480} & \textbf{*} & 200.022 & \textbf{6.161} & \textbf{0.179 \%} & 400.038 & \textbf{10.118} & \textbf{1.699 \%} & 1200.128 \\
			& POMO & 3.498 & 0.517 \% & 0.041 & 6.315 & 2.683 \% & 0.090 & 10.484 & 5.377 \% & 0.295 \\
			& Ours & 3.499 & 0.546 \% & 0.050 & 6.290 & 2.276 \% & 0.134 & 10.404 & 4.573 \% & 0.487 \\
			& Ours (ls.) & 3.481 & 0.029 \% & 0.239 & 6.171 & 0.341 \% & 0.938 & 10.126 & 1.779 \% & 3.022 \\
			& Ours (ls. + no aug.) & 3.481 & 0.029 \% & 0.196 & 6.175 & 0.407 \% & 0.807 & 10.147 & 1.990 \% & 2.671 \\
			\hline
			\multirow{6}{*}{VRPTW} & HGS-PyVRP & 7.654 & * & 200.120 & 14.545 & * & 401.586 & 24.692 & * & 1200.327 \\
			& OR-Tools & \textbf{7.654} & \textbf{*} & 200.024 & \textbf{14.724} & \textbf{1.231 \%} & 400.055 & 25.614 & 3.734 \% & 1201.444 \\
			& POMO & 7.834 & 2.352 \% & 0.040 & 15.179 & 4.359 \% & 0.095 & 26.239 & 6.265 \% & 0.345 \\
			& Ours & 7.813 & 2.077 \% & 0.052 & 15.155 & 4.194 \% & 0.143 & 26.163 & 5.957 \% & 0.550 \\
			& Ours (ls.) & 7.673 & 0.248 \% & 0.332 & 14.762 & 1.492 \% & 1.497 & \textbf{25.586} & \textbf{3.621 \%} & 4.532 \\
			& Ours (ls. + no aug.) & 7.680 & 0.340 \% & 0.289 & 14.776 & 1.588 \% & 1.407 & 25.617 & 3.746 \% & 4.072 \\
			\hline
			\multirow{6}{*}{TSP} & HGS-PyVRP & 3.829 & * & 200.127 & 5.692 & * & 401.941 & 7.757 & * & 1200.387 \\
			& OR-Tools & 3.829 & * & 200.020 & \textbf{5.692} & \textbf{*} & 400.036 & \textbf{7.768} & \textbf{0.142 \%} & 1200.023 \\
			& POMO & 3.829 & * & 0.037 & 5.697 & 0.088 \% & 0.078 & 7.828 & 0.915 \% & 0.257 \\
			& Ours & 3.829 & * & 0.045 & 5.700 & 0.141 \% & 0.116 & 7.824 & 0.864 \% & 0.421 \\
			& Ours (ls.) & 3.829 & * & 0.204 & 5.693 & 0.018 \% & 0.745 & 7.783 & 0.335 \% & 2.588 \\
			& Ours (ls. + no aug.) & 3.829 & * & 0.166 & 5.698 & 0.105 \% & 0.662 & 7.819 & 0.799 \% & 2.214 \\
			\hline
			\multirow{6}{*}{Average} & HGS-PyVRP & 4.855 & * & 200.127 & 8.677 & * & 401.508 & 14.208 & * & 1200.352 \\
			& OR-Tools & \textbf{4.855} & \textbf{*} & 200.023 & 8.733 & 0.652 \% & 400.039 & 14.594 & 2.720 \% & 1200.281 \\
			& POMO (+MD-MTA) & 4.901 & 0.948 \% & 0.055 & 8.872 & 2.251 \% & 0.115 & 14.761 & 3.890 \% & 0.327 \\
			& Ours & 4.892 & 0.774 \% & 0.056 & 8.859 & 2.101 \% & 0.140 & 14.697 & 3.444 \% & 0.498 \\
			& Ours (ls.) & 4.858 & 0.071 \% & 0.262 & \textbf{8.729} & \textbf{0.609 \%} & 1.046 & \textbf{14.472} & \textbf{1.858 \%} & 3.367 \\
			& Ours (ls. + no aug.) & 4.860 & 0.100 \% & 0.232 & 8.740 & 0.728 \% & 0.948 & 14.521 & 2.203 \% & 2.928 \\
			\hline
		\end{tabular}
	}
\end{table*}

For MDVRP instances, TuneNSearch consistently outperformed MD-MTA across all three problem sizes, even without employing the local search. When the local search is applied, the improvements became more pronounced, achieving minimal RPDs compared to PyVRP. Notably, even without x8 augmentation, the application of the local search significantly enhanced performance, allowing our model to perform better than OR-Tools and MD-MTA.

For other VRP variants, TuneNSearch generally outperformed POMO, and even OR-Tools in many cases, achieving results close to PyVRP. Similar to the MDVRP results, applying the local search led to substantial performance gains. We can also notice that the additional computational burden associated with the local search is minimal, as TuneNSearch runs for a fraction of the time required by PyVRP and OR-Tools. 

\textcolor{blue}{We also conducted an additional set of experiments using the same 100-node instances of Table~\ref{table1}, where PyVRP and OR-Tools were limited to a similar computational time budget to that of TuneNSearch.} Note that, in some variants, OR-Tools required more time than PyVRP and TuneNSearch, as this was the minimum time necessary to find a feasible solution. For TuneNSearch, we report results with the local search algorithm after inference. As shown in Table~\ref{table2}, when running under a comparable time budget, TuneNSearch outperformed PyVRP in 5 out of the 7 tested variants, achieving a lower average objective value overall. In contrast, OR-Tools did not prove to be competitive, performing significantly worse than both PyVRP and TuneNSearch in all variants. These results suggest that, in time-constrained applications, TuneNSearch is the most viable alternative.

\begin{table*}[!ht]
	\centering
	\caption{Experimental results on all VRP variants under equal computational time budget.} 
	\label{table2}
	\normalsize
	\begin{tabular}{cc|ccc}
		\hline
		& \multirow{2}{*}{Method} & \multicolumn{3}{c}{$n$ = 100} \\
		&  & Obj. & RPD & Time (m) \\
		\hline
		\multirow{3}{*}{MDVRP} & HGS-PyVRP & 11.601 & 2.256 \% & 3.505 \\
		& OR-Tools & 12.541 & 10.538 \% & 8.052 \\
		& Ours (ls.) & \textbf{11.521} & \textbf{1.151 \%} & 3.270 \\
		\hline
		\multirow{3}{*}{CVRP} & HGS-PyVRP & 16.345 & 1.371 \% & 3.501 \\
		& OR-Tools & 17.669 & 9.582 \% & 3.363 \\
		& Ours (ls.) & \textbf{16.303} & \textbf{1.110 \%} & 3.310 \\
		\hline
		\multirow{3}{*}{VRPB} & HGS-PyVRP & \textbf{13.640} & \textbf{1.155 \%} & 3.495 \\
		& OR-Tools & 14.664 & 9.026 \% & 5.361 \\
		& Ours (ls.) & 13.660 & 1.561 \% & 3.481 \\
		\hline
		\multirow{3}{*}{VRPL} & HGS-PyVRP & 16.362 & 1.388 \% & 3.501 \\
		& OR-Tools & 17.576 & 8.911 \% & 4.698 \\
		& Ours (ls.) & \textbf{16.324} & \textbf{1.153 \%} & 3.366 \\
		\hline
		\multirow{3}{*}{OVRP} & HGS-PyVRP & \textbf{10.031} & \textbf{0.824 \%} & 3.509 \\
		& OR-Tools & 10.875 & 9.307 \% & 5.371 \\
		& Ours (ls.) & 10.126 & 1.779 \% & 3.022 \\
		\hline
		\multirow{3}{*}{VRPTW} & HGS-PyVRP & 25.964 & 5.151 \% & 4.833 \\
		& OR-Tools & 26.923 & 9.035 \% & 9.408 \\
		& Ours (ls.) & \textbf{25.586} & \textbf{3.621 \%} & 4.532 \\
		\hline
		\multirow{3}{*}{TSP} & HGS-PyVRP & 7.938 & 2.333 \% & 2.818 \\
		& OR-Tools & 8.002 & 3.158 \% & 2.693 \\
		& Ours (ls.) & \textbf{7.783} & \textbf{0.335 \%} & 2.588 \\
		\hline
		\multirow{3}{*}{Average} & HGS-PyVRP & 14.554 & 2.435 \% & 3.594 \\
		& OR-Tools & 15.464 & 8.840 \% & 5.564 \\
		& Ours (ls.) & \textbf{14.472} & \textbf{1.858 \%} & 3.367 \\
		\hline
	\end{tabular}
\end{table*}

\begin{figure*}[!ht]
	\begin{center}
		\includegraphics[width=0.8\textwidth]{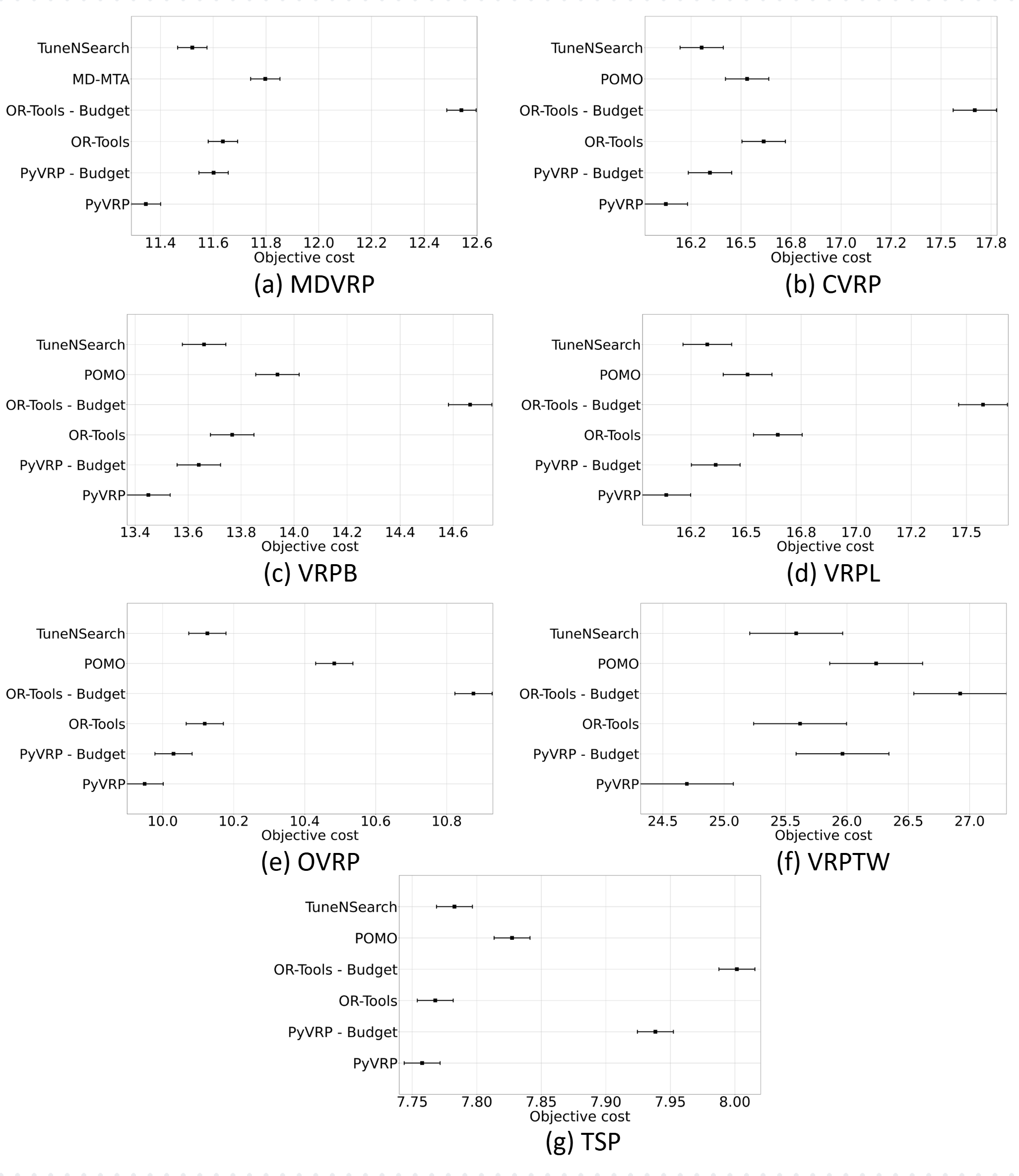}
		\caption{Means plot and 95\% confidence level Tukey's HSD intervals for different VRP variants and methods.} \label{fig4}
	\end{center}
\end{figure*}

To further support our results, we conducted a means plot, and a 95\% confidence level Tukey’s honestly significance difference (HSD) test on all 1280 generated instances for problems of sizes $n = 100$. The results, presented in Fig.~\ref{fig4}, reveal that TuneNSearch performs statistically better than POMO in nearly all variants. When compared to OR-Tools, TuneNSearch demonstrates superior performance in three variants (MDVRP, CVRP, and VRPL), with no statistically significant difference in the remaining tasks. In comparison to PyVRP, TuneNSearch shows no significant difference in two variants (CVRP and VRPL), but performs statistically worse in the others. Under a restricted time budget, TuneNSearch significantly outperforms OR-Tools across all variants. When compared to PyVRP, TuneNSearch shows no statistically significant difference in any of the variants, except for TSP, where it performs better. However, TuneNSearch exhibits a lower average objective value in 5 out of 7 variants.

\subsection{Computational results on benchmark instances}
\label{sec:5.2}

To evaluate the generalization of TuneNSearch, we tested its performance on benchmark instances for the MDVRP, VRPL, CVRP, TSP and VRPTW, as detailed in Tables~\ref{table3}, ~\ref{table4}, ~\ref{table5}, ~\ref{table6} and ~\ref{table7}. All tasks were solved greedily with x8 instance augmentation, followed by the local search algorithm (250 iterations, as we solved individual instances in this subsection). For all variants, we used models trained with instances of size $n = 100$. Additionally, we provide the solutions generated by TuneNSearch for each instance as supplementary material.

For the MDVRP, we assessed our model on Cordeau’s dataset \citep{Cordeau1997}, which includes 23 problems: instances 1-7 were proposed by \citet{Christofides1969}, instances 8-11 by \citet{Gillett1976} and instances 12-23 by \citet{Chao1993}. We compared our results to the best-known solutions (BKS) for these instances, reported by \citet{Hesam2021}. TuneNSearch was compared against the MD-MTA model, using a greedy decoding along with the enhanced depot rotation augmentation suggested in their work. 

For the VRPL, CVRP, TSP and VRPTW, we evaluated TuneNSearch using instances from CVRPLIB (Set-Golden, Set-X and Set-Solomon) \citep{Golden1998,Solomon1987,Uchoa2017} and TSPLIB \citep{Reinelt1991}. For the CVRP and TSP, we compared our results with those presented by \citet{Zhou2023}, which include POMO \citep{Kwon2020}, the adaptative multi-distribution knowledge distillation model (AMDKD-POMO) \citep{Bi2022}, the meta-learning approach (Meta-POMO) proposed by \citet{Manchanda2023}, and the omni-generalizable model (Omni-POMO) \citep{Zhou2023}. For the VRPTW, our results were compared against those reported by \citet{Zhou2024}. For the VRPL, since it is a variant not commonly evaluated on benchmark instances, we provide the results obtained by POMO.

In MDVRP benchmarks, TuneNSearch consistently outperformed the MD-MTA in all tested instances. For the VRPL, CVRP, TSP and VRPTW, TuneNSearch outperformed the other models in most instances, often by a significant margin. These results demonstrate the effectiveness of our approach and the implementation of the local search procedure. 

Furthermore, these findings underscore the strong generalization capabilities of TuneNSearch across various tasks, distributions, and problem sizes, addressing a long-standing challenge in neural-based methods. As shown in Tables~\ref{table3} and ~\ref{table4}, TuneNSearch demonstrates strong scalability as the problem size increases, maintaining computational efficiency while achieving solutions close to the BKS.

\begin{table*}[!ht]
	\centering
	\caption{Generalization on Cordeau’s benchmark instances (* represents 0.000 \% RPD).} \label{table3}
	\small
	\resizebox{\textwidth}{!}{%
		\begin{tabular}{cccc|ccccccc}
			\hline
			\multirow{2}{*}{Instance} & \multirow{2}{*}{Depots} & \multirow{2}{*}{Customers} & \multirow{2}{*}{BKS} & \multicolumn{3}{c}{Ours} & \multicolumn{3}{c}{MD-MTA} &  \\
			&  &  &  & Obj. & RPD & Time (m) & Obj. & RPD & Time (m) &  \\
			\hline
			p01 & 4 & 50 & 577 & \textbf{577} & \textbf{*} & 0.048 & 615 & 6.586 \% & 0.004 &  \\
			p02 & 4 & 50 & 474 & \textbf{480} & \textbf{1.266 \%} & 0.034 & 517 & 9.072 \% & 0.003 &  \\
			p03 & 5 & 75 & 641 & \textbf{649} & \textbf{1.248 \%} & 0.049 & 663 & 3.432 \% & 0.003 &  \\
			p04 & 2 & 100 & 1001 & \textbf{1003} & \textbf{0.200 \%} & 0.076 & 1044 & 4.296 \% & 0.004 &  \\
			p05 & 2 & 100 & 750 & \textbf{754} & \textbf{0.533 \%} & 0.069 & 785 & 4.667 \% & 0.004 &  \\
			p06 & 3 & 100 & 877 & \textbf{888} & \textbf{1.254 \%} & 0.071 & 910 & 3.763 \% & 0.004 &  \\
			p07 & 4 & 100 & 882 & \textbf{898} & \textbf{1.814 \%} & 0.073 & 929 & 5.329 \% & 0.004 &  \\
			p08 & 2 & 249 & 4372 & \textbf{4493} & \textbf{2.768 \%} & 0.283 & 4773 & 9.172 \% & 0.011 &  \\
			p09 & 3 & 249 & 3859 & \textbf{4017} & \textbf{4.094 \%} & 0.279 & 4240 & 9.873 \% & 0.012 &  \\
			p10 & 4 & 249 & 3630 & \textbf{3744} & \textbf{3.140 \%} & 0.276 & 4127 & 13.691 \% & 0.012 &  \\
			p11 & 5 & 249 & 3545 & \textbf{3632} & \textbf{2.454 \%} & 0.272 & 4034 & 13.794 \% & 0.013 &  \\
			p12 & 2 & 80 & 1319 & \textbf{1319} & \textbf{*} & 0.048 & 1390 & 5.383 \% & 0.003 &  \\
			p13 & 2 & 80 & 1319 & \textbf{1319} & \textbf{*} & 0.048 & 1390 & 5.383 \% & 0.003 &  \\
			p14 & 2 & 80 & 1360 & \textbf{1360} & \textbf{*} & 0.050 & 1390 & 2.206 \% & 0.003 &  \\
			p15 & 4 & 160 & 2505 & \textbf{2599} & \textbf{3.752 \%} & 0.126 & 2686 & 7.225 \% & 0.006 &  \\
			p16 & 4 & 160 & 2572 & \textbf{2574} & \textbf{0.078 \%} & 0.127 & 2686 & 4.432 \% & 0.006 &  \\
			p17 & 4 & 160 & 2709 & \textbf{2709} & \textbf{*} & 0.127 & 2728 & 0.701 \% & 0.006 &  \\
			p18 & 6 & 240 & 3703 & \textbf{3882} & \textbf{4.834 \%} & 0.246 & 4051 & 9.398 \% & 0.013 &  \\
			p19 & 6 & 240 & 3827 & \textbf{3832} & \textbf{0.131 \%} & 0.247 & 4051 & 5.853 \% & 0.013 &  \\
			p20 & 6 & 240 & 4058 & \textbf{4069} & \textbf{0.271 \%} & 0.245 & 4096 & 0.936 \% & 0.012 &  \\
			p21 & 9 & 360 & 5475 & \textbf{5588} & \textbf{2.064 \%} & 0.512 & 6532 & 19.306 \% & 0.039 &  \\
			p22 & 9 & 360 & 5702 & \textbf{5705} & \textbf{0.053 \%} & 0.512 & 6532 & 14.556 \% & 0.041 &  \\
			p23 & 9 & 360 & 6079 & \textbf{6129} & \textbf{0.822 \%} & 0.519 & 6532 & 7.452 \% & 0.039 &  \\
			\hline
			\multicolumn{4}{c|}{Avg. RPD} & \multicolumn{3}{c}{\textbf{1.338 \%}} & \multicolumn{3}{c}{7.239 \%} \\
			\hline
		\end{tabular}
	}
\end{table*}

\begin{table*}[!ht]
	\centering
	\caption{Generalization on VRPL instances, Set-Golden \citep{Golden1998}.} \label{table4}
	\small
	\resizebox{\textwidth}{!}{%
		\begin{tabular}{cccc|ccccccc}
			\hline
			\multirow{2}{*}{Instance} & \multirow{2}{*}{Customers} & \multirow{2}{*}{\shortstack{Distance\\Constraint}} & \multirow{2}{*}{BKS} & \multicolumn{3}{c}{Ours} & \multicolumn{4}{c}{POMO} \\
			&  &  &  & Obj. & RPD & Time (m) & Obj. & RPD & Time (m) &  \\
			\hline
			pr01 & 240 & 650 & 5623.5 & \textbf{5782.8} &\textbf{ 2.833   \%} & 0.288 & 6229.8 & 10.781 \% & 0.006 &  \\
			pr02 & 320 & 900 & 8404.6 & \textbf{8539.8} & \textbf{1.609 \%} & 0.399 & 9918.4 & 18.012 \% & 0.009 &  \\
			pr03 & 400 & 1200 & 10997.8 & \textbf{11340.5} & \textbf{3.116 \%} & 0.622 & 14067.1 & 27.908 \% & 0.013 &  \\
			pr04 & 480 & 1600 & 13588.6 & \textbf{14196.7} & \textbf{4.475 \%} & 0.861 & 17743.7 & 30.578 \% & 0.019 &  \\
			pr05 & 200 & 1800 & 6461.0 & \textbf{6582.3} & \textbf{1.877 \%} & 0.183 & 7816.9 & 20.986 \% & 0.006 &  \\
			pr06 & 280 & 1500 & 8400.3 & \textbf{8597.6} & \textbf{2.349 \%} & 0.313 & 10290.7 & 22.504 \% & 0.008 &  \\
			pr07 & 360 & 1300 & 10102.7 & \textbf{10336.8} & \textbf{2.317 \%} & 0.539 & 12941.7 & 28.101 \% & 0.011 &  \\
			pr08 & 440 & 1200 & 11635.3 & \textbf{12042.8} & \textbf{3.502 \%} & 0.767 & 16006.8 & 37.571 \% & 0.016 &  \\
			\hline
			\multicolumn{4}{c|}{Avg. RPD} & \multicolumn{3}{c}{\textbf{2.760 \%}} & \multicolumn{4}{c}{24.555 \%} \\
			\hline
		\end{tabular}
	}
\end{table*}

\begin{table*}[!ht]
	\centering
	\caption{Generalization on CVRPLIB instances, Set-X \citep{Uchoa2017}.} \label{table5}
	\small
	\resizebox{\textwidth}{!}{%
		\begin{tabular}{cc|ccccccccccc}
			\hline
			\multirow{2}{*}{Instance} & \multirow{2}{*}{BKS} & \multicolumn{2}{c}{POMO} & \multicolumn{2}{c}{AMDKD-POMO} & \multicolumn{2}{c}{Meta-POMO} & \multicolumn{2}{c}{Omni-POMO} & \multicolumn{3}{c}{Ours} \\
			&  & Obj. & RPD & Obj. & RPD & Obj. & RPD & Obj. & RPD & Obj. & \multicolumn{2}{c}{RPD} \\
			\hline
			X-n101-k25 & 27591 & 28804 & 4.396 \% & 28947 & 4.915 \% & 29647 & 7.452 \% & 29442 & 6.709 \% & \textbf{28157} & \multicolumn{2}{c}{\textbf{2.051 \%}} \\
			X-n153-k22 & 21220 & 23701 & 11.692 \% & 23179 & 9.232 \% & 23428 & 10.405 \% & 22810 & 7.493 \% & \textbf{21400} & \multicolumn{2}{c}{\textbf{0.848 \%}} \\
			X-n200-k36 & 58578 & 60983 & 4.106 \% & 61074 & 4.261 \% & 61632 & 5.214 \% & 61496 & 4.981 \% & \textbf{59322} & \multicolumn{2}{c}{\textbf{1.270 \%}} \\
			X-n251-k28 & 38684 & 40027 & 3.472 \% & 40262 & 4.079 \% & 40477 & 4.635 \% & 40059 & 3.554 \% & \textbf{39617} & \multicolumn{2}{c}{\textbf{2.412 \%}} \\
			X-n303-k21 & 21736 & 22724 & 4.545 \% & 22861 & 5.176 \% & 22661 & 4.256 \% & 22624 & 4.085 \% & \textbf{22271} & \multicolumn{2}{c}{\textbf{2.461 \%}} \\
			X-n351-k40 & 25896 & 27410 & 5.846 \% & 27431 & 5.928 \% & 27992 & 8.094 \% & 27515 & 6.252 \% & \textbf{26899} & \multicolumn{2}{c}{\textbf{3.873 \%}} \\
			X-n401-k29 & 66154 & 68435 & 3.448 \% & 68579 & 3.666 \% & 68272 & 3.202 \% & 68234 & 3.144 \% & \textbf{67406} & \multicolumn{2}{c}{\textbf{1.892 \%}} \\
			X-n459-k26 & 24139 & 26612 & 10.245 \% & 26255 & 8.766 \% & 25789 & 6.835 \% & 25706 & 6.492 \% & \textbf{25207} & \multicolumn{2}{c}{\textbf{4.424 \%}} \\
			X-n502-k39 & 69226 & 71435 & 3.191 \% & 71390 & 3.126 \% & 71209 & 2.864 \% & 70769 & 2.229 \% & \textbf{69780} & \multicolumn{2}{c}{\textbf{0.800 \%}} \\
			X-n548-k50 & 86700 & 90904 & 4.849 \% & 90890 & 4.833 \% & 90743 & 4.663 \% & 90592 & 4.489 \% & \textbf{88400} & \multicolumn{2}{c}{\textbf{1.961 \%}} \\
			X-n599-k92 & 108451 & 115894 & 6.863 \% & 115702 & 6.686 \% & 115627 & 6.617 \% & 116964 & 7.850 \% & \textbf{111898} & \multicolumn{2}{c}{\textbf{3.178 \%}} \\
			X-n655-k131 & 106780 & 110327 & 3.322 \% & 111587 & 4.502 \% & 110756 & 3.723 \% & 110096 & 3.105 \% & \textbf{107637} & \multicolumn{2}{c}{\textbf{0.803 \%}} \\
			X-n701-k44 & 81923 & 86933 & 6.115 \% & 88166 & 7.621 \% & 86605 & 5.715 \% & 86005 & 4.983 \% & \textbf{84894} & \multicolumn{2}{c}{\textbf{3.627 \%}} \\
			X-n749-k98 & 77269 & 83294 & 7.797 \% & 83934 & 8.626 \% & 84406 & 9.237 \% & 83893 & 8.573 \% & \textbf{80278} & \multicolumn{2}{c}{\textbf{3.894 \%}} \\
			X-n801-k40 & 73311 & 80584 & 9.921 \% & 80897 & 10.348 \% & 79077 & 7.865 \% & 78171 & 6.630 \% & \textbf{75870} & \multicolumn{2}{c}{\textbf{3.491 \%}} \\
			X-n856-k95 & 88965 & 96398 & 8.355 \% & 95809 & 7.693 \% & 95801 & 7.684 \% & 96739 & 8.748 \% & \textbf{90418} & \multicolumn{2}{c}{\textbf{1.633 \%}} \\
			X-n895-k37 & 53860 & 61604 & 14.378 \% & 62316 & 15.700 \% & 59778 & 10.988 \% & 58947 & 9.445 \% & \textbf{56456} & \multicolumn{2}{c}{\textbf{4.820 \%}} \\
			X-n957-k87 & 85465 & 93221 & 9.075 \% & 93995 & 9.981 \% & 92647 & 8.403 \% & 92011 & 7.659 \% & \textbf{87563} & \multicolumn{2}{c}{\textbf{2.455 \%}} \\
			X-n1001-k43 & 72355 & 82046 & 13.394 \% & 82855 & 14.512 \% & 79347 & 9.663 \% & 78955 & 9.122 \% & \textbf{76447} & \multicolumn{2}{c}{\textbf{5.655 \%}} \\
			\hline
			\multicolumn{2}{c|}{Avg. RPD} & \multicolumn{2}{c}{7.106 \%} & \multicolumn{2}{c}{7.353 \%} & \multicolumn{2}{c}{6.711 \%} & \multicolumn{2}{c}{6.081 \%} & \multicolumn{3}{c}{\textbf{2.713 \%}} \\
			\hline
		\end{tabular}
	}
\end{table*}

\begin{table*}[!ht]
	\centering
	\caption{Generalization on TSPLIB instances \citep{Reinelt1991}.} \label{table6}
	\small
	\resizebox{\textwidth}{!}{%
		\begin{tabular}{cc|ccccccccccccccc}
			\hline
			\multirow{2}{*}{Instance} & \multirow{2}{*}{BKS} & \multicolumn{3}{c}{POMO} & \multicolumn{3}{c}{AMDKD-POMO} & \multicolumn{3}{c}{Meta-POMO} & \multicolumn{3}{c}{Omni-POMO} & \multicolumn{3}{c}{Ours} \\
			&  & Obj. & \multicolumn{2}{c}{RPD} & Obj. & \multicolumn{2}{c}{RPD} & Obj. & \multicolumn{2}{c}{RPD} & Obj. & \multicolumn{2}{c}{RPD} & Obj. & \multicolumn{2}{c}{RPD} \\
			\hline
			kroA100 & 21282 & \textbf{21282} & \multicolumn{2}{c}{\textbf{*}} & 21360 & \multicolumn{2}{c}{0.366 \%} & 21308 & \multicolumn{2}{c}{0.122 \%} & 21305 & \multicolumn{2}{c}{0.108 \%} & 21306 & \multicolumn{2}{c}{0.113 \%} \\
			kroA150 & 26524 & \textbf{26823} & \multicolumn{2}{c}{\textbf{1.127 \%}} & 26997 & \multicolumn{2}{c}{1.783 \%} & 26852 & \multicolumn{2}{c}{1.237 \%} & 26873 & \multicolumn{2}{c}{1.316 \%} & 26875 & \multicolumn{2}{c}{1.323 \%} \\
			kroA200 & 29368 & \textbf{29745} & \multicolumn{2}{c}{\textbf{1.284 \%}} & 30196 & \multicolumn{2}{c}{2.819 \%} & 29749 & \multicolumn{2}{c}{1.297 \%} & 29823 & \multicolumn{2}{c}{1.549 \%} & 29770 & \multicolumn{2}{c}{1.369 \%} \\
			kroB200 & 29437 & 30060 & \multicolumn{2}{c}{2.116 \%} & 30188 & \multicolumn{2}{c}{2.551 \%} & 29896 & \multicolumn{2}{c}{1.559 \%} & 29814 & \multicolumn{2}{c}{1.281 \%} & \textbf{29800} & \multicolumn{2}{c}{\textbf{1.233 \%}} \\
			ts225 & 126643 & 131208 & \multicolumn{2}{c}{3.605 \%} & 128210 & \multicolumn{2}{c}{1.237 \%} & 131877 & \multicolumn{2}{c}{4.133 \%} & 128770 & \multicolumn{2}{c}{1.679 \%} & \textbf{127763} & \multicolumn{2}{c}{\textbf{0.884 \%}} \\
			tsp225 & 3916 & 4040 & \multicolumn{2}{c}{3.166 \%} & 4074 & \multicolumn{2}{c}{4.035 \%} & 4047 & \multicolumn{2}{c}{3.345 \%} & 4008 & \multicolumn{2}{c}{2.349 \%} & \textbf{3976} & \multicolumn{2}{c}{\textbf{1.532 \%}} \\
			pr226 & 80369 & 81509 & \multicolumn{2}{c}{1.418 \%} & 82430 & \multicolumn{2}{c}{2.564 \%} & 81968 & \multicolumn{2}{c}{1.990\%} & 81839 & \multicolumn{2}{c}{1.829 \%} & \textbf{80735} & \multicolumn{2}{c}{\textbf{0.455 \%}} \\
			pr264 & 49135 & 50513 & \multicolumn{2}{c}{2.804 \%} & 51656 & \multicolumn{2}{c}{5.131 \%} & 50065 & \multicolumn{2}{c}{1.893 \%} & 50649 & \multicolumn{2}{c}{3.081 \%} & \textbf{49653} & \multicolumn{2}{c}{\textbf{1.054 \%}} \\
			a280 & 2579 & 2714 & \multicolumn{2}{c}{5.234 \%} & 2773 & \multicolumn{2}{c}{7.522 \%} & 2703 & \multicolumn{2}{c}{4.808 \%} & 2695 & \multicolumn{2}{c}{4.498 \%} & \textbf{2632} & \multicolumn{2}{c}{\textbf{2.055 \%}} \\
			pr299 & 48191 & 50571 & \multicolumn{2}{c}{4.939 \%} & 51270 & \multicolumn{2}{c}{6.389 \%} & 49773 & \multicolumn{2}{c}{3.283 \%} & 49348 & \multicolumn{2}{c}{2.401 \%} & \textbf{48833} & \multicolumn{2}{c}{\textbf{1.332 \%}} \\
			lin318 & 42029 & 44011 & \multicolumn{2}{c}{4.716 \%} & 44154 & \multicolumn{2}{c}{5.056 \%} & 43807 & \multicolumn{2}{c}{4.230 \%} & 43828 & \multicolumn{2}{c}{4.280 \%} & \textbf{43022} & \multicolumn{2}{c}{\textbf{2.363 \%}} \\
			rd400 & 15281 & 16254 & \multicolumn{2}{c}{6.367 \%} & 16610 & \multicolumn{2}{c}{8.697 \%} & 16153 & \multicolumn{2}{c}{5.706 \%} & 15948 & \multicolumn{2}{c}{4.365 \%} & \textbf{15794} & \multicolumn{2}{c}{\textbf{3.357 \%}} \\
			fl417 & 11861 & 12940 & \multicolumn{2}{c}{9.097 \%} & 13129 & \multicolumn{2}{c}{10.690 \%} & 12849 & \multicolumn{2}{c}{8.330 \%} & 12683 & \multicolumn{2}{c}{6.930 \%} & \textbf{11944} & \multicolumn{2}{c}{\textbf{0.700 \%}} \\
			pr439 & 107217 & 115651 & \multicolumn{2}{c}{7.866 \%} & 117872 & \multicolumn{2}{c}{9.938 \%} & 114872 & \multicolumn{2}{c}{7.140 \%} & 114487 & \multicolumn{2}{c}{6.781 \%} & \textbf{111502} & \multicolumn{2}{c}{\textbf{3.997 \%}} \\
			pcb442 & 50778 & 55273 & \multicolumn{2}{c}{8.852 \%} & 56225 & \multicolumn{2}{c}{10.727 \%} & 55507 & \multicolumn{2}{c}{9.313 \%} & 54531 & \multicolumn{2}{c}{7.391 \%} & \textbf{52635} & \multicolumn{2}{c}{\textbf{3.657 \%}} \\
			d493 & 35002 & 38388 & \multicolumn{2}{c}{9.674 \%} & 38400 & \multicolumn{2}{c}{9.708 \%} & 38641 & \multicolumn{2}{c}{10.396 \%} & 38169 & \multicolumn{2}{c}{9.048 \%} & \textbf{36975} & \multicolumn{2}{c}{\textbf{5.637 \%}} \\
			u574 & 36905 & 41574 & \multicolumn{2}{c}{12.651 \%} & 41426 & \multicolumn{2}{c}{12.250 \%} & 41418 & \multicolumn{2}{c}{12.229 \%} & 40515 & \multicolumn{2}{c}{9.782 \%} & \textbf{38918} & \multicolumn{2}{c}{\textbf{5.454 \%}} \\
			rat575 & 6773 & 7617 & \multicolumn{2}{c}{12.461 \%} & 7707 & \multicolumn{2}{c}{13.790 \%} & 7620 & \multicolumn{2}{c}{12.505 \%} & 7658 & \multicolumn{2}{c}{13.067 \%} & \textbf{7197} & \multicolumn{2}{c}{\textbf{6.260 \%}} \\
			p654 & 34643 & 38556 & \multicolumn{2}{c}{11.295 \%} & 39327 & \multicolumn{2}{c}{13.521 \%} & 38307 & \multicolumn{2}{c}{10.576 \%} & 37488 & \multicolumn{2}{c}{8.212 \%} & \textbf{35265} & \multicolumn{2}{c}{\textbf{1.795 \%}} \\
			d657 & 48912 & 55133 & \multicolumn{2}{c}{12.719 \%} & 55143 & \multicolumn{2}{c}{12.739 \%} & 54715 & \multicolumn{2}{c}{11.864 \%} & 54346 & \multicolumn{2}{c}{11.110 \%} & \textbf{51510} & \multicolumn{2}{c}{\textbf{5.312 \%}} \\
			u724 & 41910 & 48855 & \multicolumn{2}{c}{16.571 \%} & 48738 & \multicolumn{2}{c}{16.292 \%} & 48272 & \multicolumn{2}{c}{15.180 \%} & 48026 & \multicolumn{2}{c}{14.593 \%} & \textbf{43886} & \multicolumn{2}{c}{\textbf{4.715 \%}} \\
			rat783 & 8806 & 10401 & \multicolumn{2}{c}{18.113 \%} & 10338 & \multicolumn{2}{c}{17.397 \%} & 10228 & \multicolumn{2}{c}{16.148 \%} & 10300 & \multicolumn{2}{c}{16.966 \%} & \textbf{9409} & \multicolumn{2}{c}{\textbf{6.848 \%}} \\
			pr1002 & 259045 & 310855 & \multicolumn{2}{c}{20.000 \%} & 312299 & \multicolumn{2}{c}{20.558 \%} & 308281 & \multicolumn{2}{c}{19.007 \%} & 305777 & \multicolumn{2}{c}{18.040 \%} & \textbf{278844} & \multicolumn{2}{c}{\textbf{7.643 \%}} \\
			\hline
			\multicolumn{2}{c|}{Avg. RPD} & \multicolumn{3}{c}{7.264 \%} & \multicolumn{3}{c}{8.511 \%} & \multicolumn{3}{c}{7.230 \%} & \multicolumn{3}{c}{6.550 \%} & \multicolumn{3}{c}{\textbf{3.004 \%}} \\
			\hline
		\end{tabular}
	}
\end{table*}

\begin{table*}[!ht]
	\centering
	\caption{Generalization on VRPTW instances, Set-Solomon \citep{Solomon1987}.} \label{table7}
	\small
	\resizebox{\textwidth}{!}{%
		\begin{tabular}{cc|ccccccccccl}
			\hline
			\multirow{2}{*}{Instance} & \multirow{2}{*}{BKS} & \multicolumn{2}{c}{POMO} & \multicolumn{2}{c}{POMO-MTL} & \multicolumn{2}{c}{MVMoE/4E} & \multicolumn{2}{c}{MVMoE/4E-L} & \multicolumn{3}{c}{Ours} \\
			&  & Obj. & RPD & Obj. & RPD & Obj. & RPD & Obj. & RPD & Obj. & \multicolumn{2}{c}{RPD} \\
			\hline
			R101 & 1637.7 & 1805.6 & 10.252 \% & 1821.2 & 11.205 \% & 1798.1 & 9.794 \% & 1730.1 & 5.641 \% & \textbf{1644.2} & \multicolumn{2}{c}{\textbf{0.400 \%}} \\
			R102 & 1466.6 & 1556.7 & 6.143 \% & 1596.0 & 8.823 \% & 1572.0 & 7.187 \% & 1574.3 & 7.345 \% & \textbf{1493.7} & \multicolumn{2}{c}{\textbf{1.848 \%}} \\
			R103 & 1208.7 & 1341.4 & 10.979 \% & 1327.3 & 9.812 \% & 1328.2 & 9.887 \% & 1359.4 & 12.470 \% & \textbf{1223.5} & \multicolumn{2}{c}{\textbf{1.224 \%}} \\
			R104 & 971.5 & 1118.6 & 15.142 \% & 1120.7 & 15.358 \% & 1124.8 & 15.780 \% & 1098.8 & 13.100 \% & \textbf{977.8} & \multicolumn{2}{c}{\textbf{0.648 \%}} \\
			R105 & 1355.3 & 1506.4 & 11.149 \% & 1514.6 & 11.754 \% & 1479.4 & 9.157 \% & 1456.0 & 7.433 \% & \textbf{1364.5} & \multicolumn{2}{c}{\textbf{0.679 \%}} \\
			R106 & 1234.6 & 1365.2 & 10.578 \% & 1380.5 & 11.818 \% & 1362.4 & 10.352 \% & 1353.5 & 9.627 \% & \textbf{1249.6} & \multicolumn{2}{c}{\textbf{1.215 \%}} \\
			R107 & 1064.6 & 1214.2 & 14.052 \% & 1209.3 & 13.592 \% & 1181.1 & 11.037 \% & 1196.5 & 12.391 \% & \textbf{1099.3} & \multicolumn{2}{c}{\textbf{3.259 \%}} \\
			R108 & 932.1 & 1058.9 & 13.604 \% & 1061.8 & 13.915 \% & 1023.2 & 9.774 \% & 1039.1 & 11.481 \% & \textbf{961.2} & \multicolumn{2}{c}{\textbf{3.122 \%}} \\
			R109 & 1146.9 & 1249.0 & 8.902 \% & 1265.7 & 10.358 \% & 1255.6 & 9.478 \% & 1224.3 & 6.750 \% & \textbf{1158.5} & \multicolumn{2}{c}{\textbf{1.011 \%}} \\
			R110 & 1068.0 & 1180.4 & 10.524 \% & 1171.4 & 9.682 \% & 1185.7 & 11.021 \% & 1160.2 & 8.635 \% & \textbf{1087.9} & \multicolumn{2}{c}{\textbf{1.863 \%}} \\
			R111 & 1048.7 & 1177.2 & 12.253 \% & 1211.5 & 15.524 \% & 1176.1 & 12.148 \% & 1197.8 & 14.220 \% & \textbf{1060.1} & \multicolumn{2}{c}{\textbf{1.087 \%}} \\
			R112 & 948.6 & 1063.1 & 12.070 \% & 1057.0 & 11.427 \% & 1045.2 & 10.183 \% & 1044.2 & 10.082 \% & \textbf{961.8} & \multicolumn{2}{c}{\textbf{1.391 \%}} \\
			RC101 & 1619.8 & 2643.0 & 63.168 \% & 1833.3 & 13.181 \% & 1774.4 & 9.544 \% & 1749.2 & 7.988\% & \textbf{1639.8} & \multicolumn{2}{c}{\textbf{1.235 \%}} \\
			RC102 & 1457.4 & 1534.8 & 5.311 \% & 1546.1 & 6.086 \% & 1544.5 & 5.976 \% & 1556.1 & 6.771 \% & \textbf{1477.2} & \multicolumn{2}{c}{\textbf{1.359 \%}} \\
			RC103 & 1258.0 & 1407.5 & 11.884 \% & 1396.2 & 10.986 \% & 1402.5 & 11.486 \% & 1415.3 & 12.502 \% & \textbf{1279.1} & \multicolumn{2}{c}{\textbf{1.677 \%}} \\
			RC104 & 1132.3 & 1261.8 & 11.437 \% & 1271.7 & 12.311 \% & 1265.4 & 11.755 \% & 1264.2 & 11.649 \% & \textbf{1163.4} & \multicolumn{2}{c}{\textbf{2.747 \%}} \\
			RC105 & 1513.7 & 1612.9 & 6.553 \% & 1644.9 & 8.668 \% & 1635.5 & 8.047 \% & 1619.4 & 6.980 \% & \textbf{1549.2} & \multicolumn{2}{c}{\textbf{2.345 \%}} \\
			RC106 & 1372.7 & 1539.3 & 12.137 \% & 1552.8 & 13.120 \% & 1505.0 & 9.638 \% & 1509.5 & 9.968 \% & \textbf{1391.0} & \multicolumn{2}{c}{\textbf{1.333 \%}} \\
			RC107 & 1207.8 & 1347.7 & 11.583 \% & 1384.8 & 14.655 \% & 1351.6 & 11.906 \% & 1324.1 & 9.625 \% & \textbf{1214.9} & \multicolumn{2}{c}{\textbf{0.588 \%}} \\
			RC108 & 1114.2 & 1305.5 & 17.169 \% & 1274.4 & 14.378 \% & 1254.2 & 12.565 \% & 1247.2 & 11.939 \% & \textbf{1134.3} & \multicolumn{2}{c}{\textbf{1.804 \%}} \\
			RC201 & 1261.8 & 2045.6 & 62.118 \% & 1761.1 & 39.570 \% & 1577.3 & 25.004 \% & 1517.8 & 20.285 \% & \textbf{1280.9} & \multicolumn{2}{c}{\textbf{1.514 \%}} \\
			RC202 & 1092.3 & 1805.1 & 65.257 \% & 1486.2 & 36.062 \% & 1616.5 & 47.990 \% & 1480.3 & 35.520 \% & \textbf{1106.6} & \multicolumn{2}{c}{\textbf{1.309 \%}} \\
			RC203 & 923.7 & 1470.4 & 59.186 \% & 1360.4 & 47.277 \% & 1473.5 & 59.521 \% & 1479.6 & 60.182 \% & \textbf{940.4} & \multicolumn{2}{c}{\textbf{1.808 \%}} \\
			RC204 & 783.5 & 1323.9 & 68.973 \% & 1331.7 & 69.968 \% & 1286.6 & 64.212 \% & 1232.8 & 57.342 \% & \textbf{791.4} & \multicolumn{2}{c}{\textbf{1.008 \%}} \\
			RC205 & 1154.0 & 1568.4 & 35.910 \% & 1539.2 & 33.380 \% & 1537.7 & 33.250 \% & 1440.8 & 24.850 \% & \textbf{1171.1} & \multicolumn{2}{c}{\textbf{1.482 \%}} \\
			RC206 & 1051.1 & 1707.5 & 62.449 \% & 1472.6 & 40.101 \% & 1468.9 & 39.749 \% & 1394.5 & 32.671 \% & \textbf{1086.3} & \multicolumn{2}{c}{\textbf{3.349 \%}} \\
			RC207 & 962.9 & 1567.2 & 62.758 \% & 1375.7 & 42.870 \% & 1442.0 & 49.756 \% & 1346.4 & 39.831 \% & \textbf{981.9} & \multicolumn{2}{c}{\textbf{1.973 \%}} \\
			RC208 & 776.1 & 1505.4 & 93.970 \% & 1185.6 & 52.764 \% & 1107.4 & 42.688 \% & 1167.5 & 50.437 \% & \textbf{782.2} & \multicolumn{2}{c}{\textbf{0.786 \%}} \\
			\hline
			\multicolumn{2}{c|}{Avg. RPD} & \multicolumn{2}{c}{28.054 \%} & \multicolumn{2}{c}{21.380 \%} & \multicolumn{2}{c}{20.317 \%} & \multicolumn{2}{c}{18.490 \%} & \multicolumn{3}{c}{\textbf{1.574 \%}} \\
			\hline
		\end{tabular}
	}
\end{table*}

\subsection{Why pre-train on the MDVRP?}
\label{sec:5.3}

We argue that one of the main contributions of this paper is that pre-training the model on the MDVRP can lead to a better generalization across different single- and multi-depot tasks. To assess the effectiveness of this approach, we compared TuneNSearch with a model pre-trained on the CVRP. We evaluate both models on their zero-shot generalization performance to both single-depot and multi-depot variants. \textcolor{blue}{Zero-shot generalization refers to the model’s ability to perform well on new VRP variants it has not seen during training, without any additional fine-tuning.} For the multi-depot variants, we incorporated the constraints outlined in Section~\ref{sec:3.3} into the classic MDVRP, resulting in the following variants: MDVRP with backhauls (MDVRPB), MDVRP with duration limits (MDVRPL), MDVRP with open routes (MDOVRP), and MDVRP with time windows (MDVRPTW). 

Like TuneNSearch, the CVRP-based model was trained for 100 epochs, using the same deep neural network architecture. Table~\ref{table8} presents the inference results for both models in terms of the objective value and the performance deviation of TuneNSearch relative to the CVRP model. The evaluation was conducted on 1280 randomly generated instances for each VRP variant. In both cases, inference was performed greedily with instance augmentation. In these experiments, we did not apply the local search algorithm.

\textcolor{blue}{Results show that TuneNSearch performs similarly to the CVRP-based model on single-depot variants, with a negligible performance deviation between both methods.} However, on multi-depot variants, TuneNSearch performs significantly better on all instance sizes. The performance deviation becomes especially more pronounced as problem size increases: while the model pre-trained on the CVRP performs reasonably well on small instances (with 20 nodes), its performance degrades sharply on larger problems.

\begin{table*}[!ht]
	\centering
	\caption{Zero-shot generalization comparison on 1280 randomly generated instances: Pre-training on MDVRP vs. CVRP.} \label{table8}
	\footnotesize
	\resizebox{\textwidth}{!}{%
		\begin{tabular}{cc|cc|cc|cc}
			\hline
			\multirow{2}{*}{} & \multirow{2}{*}{Method} & \multicolumn{2}{c|}{$n$ = 20} & \multicolumn{2}{c|}{$n$ = 50} & \multicolumn{2}{c}{$n$ = 100} \\
			&  & Obj. & Performance deviation & Obj. & Performance deviation & Obj. & Performance deviation \\
			\hline
			\multirow{2}{*}{VRPB} & Ours & 4.665 & \multirow{2}{*}{0.172 \%} & 8.752 & \multirow{2}{*}{0.436 \%} & 15.040 & \multirow{2}{*}{0.220 \%} \\
			& CVRP pre-trained & \textbf{4.657} &  & \textbf{8.714} &  & \textbf{15.007} &  \\
			\hline
			\multirow{2}{*}{VRPL} & Ours & 5.036 & \multirow{2}{*}{0.219 \%} & 9.550 & \multirow{2}{*}{0.569 \%} & 16.597 & \multirow{2}{*}{0.533 \%} \\
			& CVRP pre-trained & \textbf{5.025} & & \textbf{9.496} &  & \textbf{16.509} &  \\
			\hline
			\multirow{2}{*}{OVRP} & Ours & \textbf{3.879} & \multirow{2}{*}{-0.589 \%} & 7.257 & \multirow{2}{*}{0.152 \%} & 12.223 & \multirow{2}{*}{0.767 \%} \\
			& CVRP pre-trained & 3.902 &  & \textbf{7.246} &  & \textbf{12.130} &  \\
			\hline
			\multirow{2}{*}{VRPTW} & Ours & \textbf{8.652} & \multirow{2}{*}{-1.075 \%} & 18.340 & \multirow{2}{*}{1.164 \%} & \textbf{31.468} & \multirow{2}{*}{-0.691 \%} \\
			& CVRP pre-trained & 8.746 &  & \textbf{18.129} &  & 31.687 &  \\
			\hline
			\multirow{2}{*}{\shortstack{Average\\(Single-depot)}}& Ours & \textbf{5.558} & \multirow{2}{*}{-0.430 \%} & 10.975 & \multirow{2}{*}{0.725 \%} & \textbf{18.832} & \multirow{2}{*}{-0.005 \%} \\
			& CVRP pre-trained & 5.582 &  & \textbf{10.896} &  & 18.833 &  \\
			\hline
			\multirow{2}{*}{MDVRPB} & Ours & \textbf{4.343} & \multirow{2}{*}{-11.494 \%} & \textbf{7.203} & \multirow{2}{*}{-22.490 \%} & \textbf{11.003} & \multirow{2}{*}{-49.530 \%} \\
			& CVRP pre-trained & 4.907 &  & 9.293 &  & 21.801 &  \\
			\hline
			\multirow{2}{*}{MDVRPL} & Ours & \textbf{4.462} & \multirow{2}{*}{-11.169 \%} & \textbf{7.514} & \multirow{2}{*}{-22.695 \%} & \textbf{11.540} & \multirow{2}{*}{-46.596 \%} \\
			& CVRP pre-trained & 5.023 & & 9.720 &  & 21.609 &  \\
			\hline
			\multirow{2}{*}{MDOVRP} & Ours & \textbf{3.613} & \multirow{2}{*}{-8.508 \%} & \textbf{6.046} & \multirow{2}{*}{-20.279 \%} & \textbf{9.253} & \multirow{2}{*}{-40.518 \%} \\
			& CVRP pre-trained & 3.949 &  & 7.584 &  & 15.556 &  \\
			\hline
			\multirow{2}{*}{MDVRPTW} & Ours & \textbf{8.795} & \multirow{2}{*}{-12.981 \%} & \textbf{17.054} & \multirow{2}{*}{-22.256 \%} & \textbf{27.310} & \multirow{2}{*}{-43.012 \%} \\
			& CVRP pre-trained & 10.107 &  & 21.936 &  & 47.922 &  \\
			\hline
			\multirow{2}{*}{\shortstack{Average\\(Multi-depot)}}& Ours & \textbf{5.303} & \multirow{2}{*}{-11.558 \%} & \textbf{9.454} & \multirow{2}{*}{-22.080 \%} & \textbf{14.776} & \multirow{2}{*}{-44.705 \%} \\
			& CVRP pre-trained & 5.996 &  & 12.133 &  & 26.722 &  \\
			\hline
		\end{tabular}
	}
\end{table*}

\subsection{Local search effectiveness assessment}
\label{sec:5.4}

\begin{figure*}[!ht]
	\begin{center}
		\includegraphics[width=\textwidth]{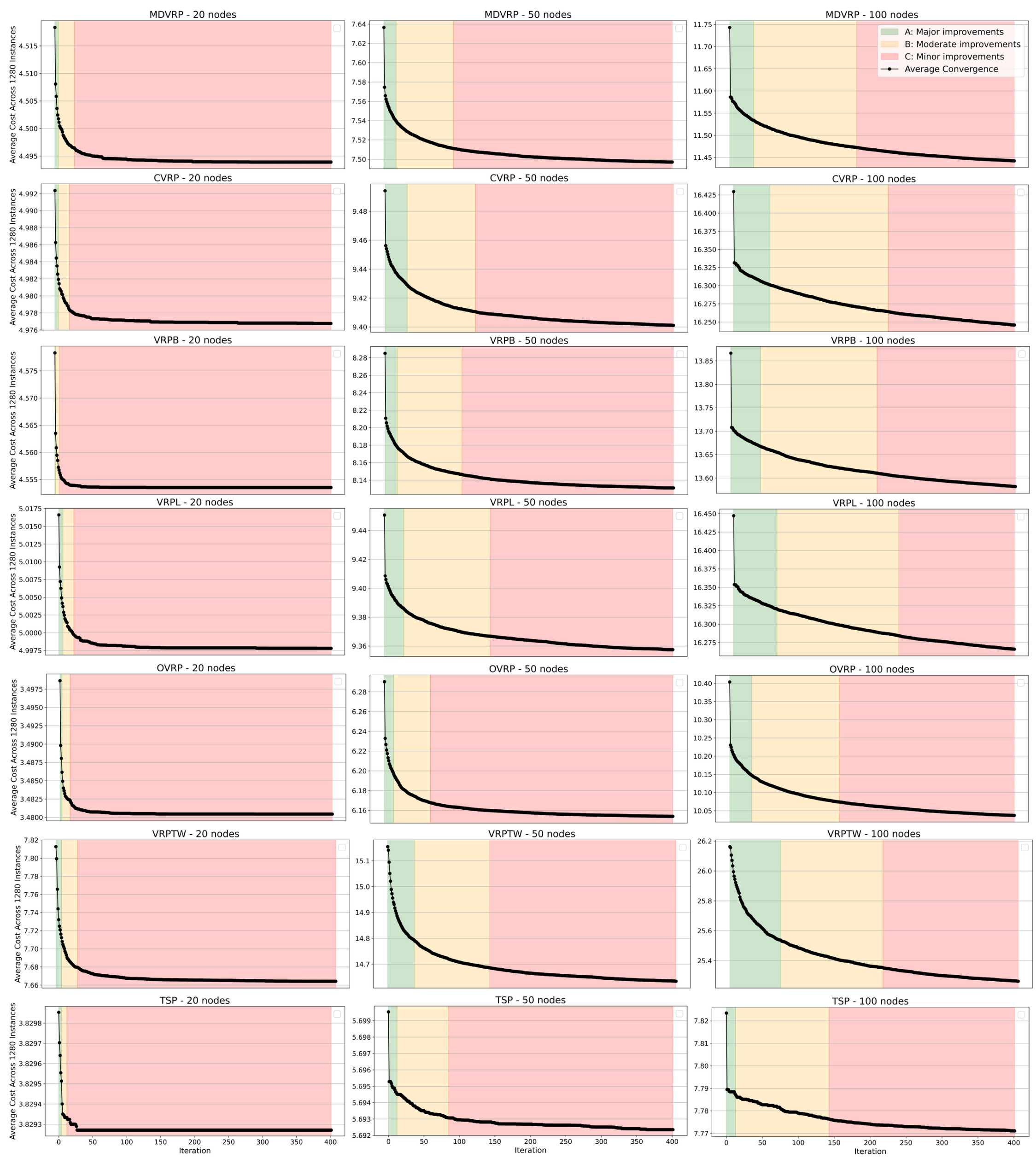}
		\caption{Local search convergence of all variants across 1280 randomly generated instances.} \label{fig5}
	\end{center}
\end{figure*}

\textcolor{blue}{To evaluate the impact of the local search algorithm, we conducted a sensitivity analysis on its convergence behavior over successive iterations, shown in Fig.~\ref{fig5}. This analysis allowed us to examine how solution quality evolves with increasing iterations and how quickly the algorithm converges. Specifically, we tracked the evolution of the best solution found by our method over 400 iterations of the local search for all tested problem variants, on instances with 20, 50 and 100 nodes. For these experiments, we used the same models as in Section~\ref{sec:5.1} (with label \textit{Ours (ls.)} in Table~\ref{table1}). For each variant-size combination, the analysis included 1280 randomly generated instances, and the plots report the average of the best solutions obtained by the model up to each iteration. To further understand the convergence behavior of our method, we performed an ABC analysis for each variant-size pair, represented by the green, yellow, and red shaded areas in each subplot of Fig.~\ref{fig5}. The "A" group (green) shows the early iterations, during which the algorithm achieves majority of its improvement (specifically, 70\% of the total improvement relative to the final solutions at iteration 400). The "B" group (yellow) covers the iterations during which the algorithm reaches a 90\% of the final improvement, while the "C" group (red) represents the final 10\% of improvement.}

\textcolor{blue}{Overall, we can identify different patterns through this sensitivity analysis. On smaller problems, across all variants tested, in most cases the algorithm reached the "B" group before the 10\textsuperscript{th} iteration, showing a very fast convergence. For problems with 50 nodes, this happened, in general, between iterations 20 and 30, while for problems with 100 nodes, about 70 iterations were needed, in the worst case. We can also notice that some variants displayed a much smoother convergence than others. For example, tasks such as VRPTW and VRPL demonstrated a higher sensitivity to changes in the number of iterations. In contrast, simpler variants like the TSP exhibited much lower sensitivity, with a very rapid convergence even on 100-node instances.}

\textcolor{blue}{Overall, we found that 50 iterations offer a good trade-off between solution quality and computational performance. Beyond this point, further iterations yield  marginal improvements, with the objective function improving at a much slower rate while imposing more computational time. Since TuneNSearch was designed intended for large-scale, time-sensitive applications where scalability is critical, extending the local search for too many iterations would undermine the efficiency benefits introduced by the reinforcement learning component, which goes against our primary design goal.}

\textcolor{blue}{Having analyzed the convergence behavior of the local search algorithm, we now turn our attention to the role of solution initialization. In particular, we examine whether better initial solutions significantly affect the final solution quality. TuneNSearch benefits from a neural-based model to generate high-quality solutions, starting from more promising regions of the search space and accelerating convergence. Therefore, to assess the effectiveness of this approach, we compare it with two other solution initialization strategies. The first initialization strategy (Random) begins with an empty solution, and constructs routes sequentially. Customer nodes are assigned to the current route in a random order, as long as the vehicle has enough capacity to meet their demand. Once a route can no longer accommodate additional customers, a new route is started, and the process continues until all customers are assigned. The resulting solutions are then improved through the same local search procedure as in our method. In the second initialization strategy (Greedy), routes are also built sequentially. However, instead of assigning customers randomly, each customer is chosen based on the nearest neighbor method, that is, the closest unvisited customer is chosen, provided the vehicle has enough capacity.}

\textcolor{blue}{As in previous experiments, each approach was tested on 1280 randomly generated instances for the different variants solved in Section~\ref{sec:5.1}. The experiments included problems with 20, 50, 100 and 200 nodes. We evaluated the performance of each approach using three different metrics: Obj., the average objective function across all 1280 instances; Win rate \%, the percentage of instances in which each of the approaches outperformed the others; Time (m), the total computational time (in minutes) required to solve all 1280 instances. Note that the Win rate \% values do not always sum to 100\% since all three approaches may achieve identical solutions on the same instance. All results are shown in Table~\ref{table9}.}

\textcolor{blue}{Overall, our neural-based initialization consistently outperformed the other two strategies across almost all instance sizes and variants. We can also notice that the effectiveness of TuneNSearch becomes increasingly pronounced as instance size grows. For smaller problems, all strategies perform similarly because the search space is limited, allowing the local search to refine even lower-quality initial solutions effectively. However, on problems with 100 and 200 nodes, our method outperformed the other in the majority of instances, achieving a Win (\%) of around 70-80\% compared to the other initialization strategies. This suggests that the neural-based initialization allows the local search algorithm to significantly bypass the initial effort required to get close to promising regions of the solution space.}

\begin{table*}[!ht]
	\centering
	\caption{\textcolor{blue}{Comparison of neural-based, random and greedy solution initialization strategies on 1280 randomly generated instances.}}
	\label{table9}
	\small
	\resizebox{\textwidth}{!}{%
		\textcolor{blue}{
		\begin{tabular}{cc|ccc|ccc|ccc|ccc}
			\hline
			& \multirow{2}{*}{Method}
			& \multicolumn{3}{c|}{$n = 20$} 
			& \multicolumn{3}{c|}{$n = 50$} 
			& \multicolumn{3}{c|}{$n = 100$} 
			& \multicolumn{3}{c}{$n = 200$} \\
			&  & Obj. & Win (\%) & Time (m) & Obj. & Win (\%) & Time & Obj. & Win \% & Time & Obj. & Win (\%) & Time \\
			\hline
			\multirow{3}{*}{MDVRP} 
			& Ours   & \textbf{4.504} & \textbf{4.14\%} & 0.271 &\textbf{ 7.525} & \textbf{34.69\%} & 1.059 & \textbf{11.485} & \textbf{75.08\%} & 2.989 & \textbf{19.567} & \textbf{86.17\%} & 11.381 \\
			& Random & 4.508 & 0.94\% & 0.266 & 7.555 & 14.30\% & 1.237 & 11.640 & 12.19\% & 3.388 & 19.910 & 5.55\% & 11.492 \\
			& Greedy & 4.507 & 1.02\% & 0.251 & 7.552 & 15.55\% & 1.249 & 11.635 & 12.73\% & 3.412 & 19.887 & 8.28\% & 11.648 \\
			\hline
			\multirow{3}{*}{CVRP} 
			& Ours   & \textbf{4.978} & \textbf{1.88\%} & 0.251 & \textbf{9.398} & \textbf{35.23\%} & 1.024 & \textbf{16.235} & \textbf{84.77\%} & 2.926 & \textbf{29.509} & \textbf{88.36\%} & 11.385 \\
			& Random & 5.011 & 0.08\% & 0.248 & 9.419 & 14.77\% & 1.274 & 16.418 & 8.83\% & 3.563 & 29.871 & 5.23\% & 12.231 \\
			& Greedy & 4.989 & 0.62\% & 0.251 & 9.419 & 15.70\% & 1.281 & 16.425 & 5.94\% & 3.591 & 29.861 & 6.09\% & 12.468 \\
			\hline
			\multirow{3}{*}{VRPB} 
			& Ours   & 4.606 & \textbf{0.55\%} & 0.257 & \textbf{8.383} & \textbf{33.83\%} & 1.012 & \textbf{14.176} & \textbf{71.64\%} & 2.898 & \textbf{25.209} & \textbf{77.89\%} & 10.884 \\
			& Random & 4.606 & 0.23\% & 0.248 & 8.400 & 20.08\% & 1.262 & 14.305 & 13.44\% & 3.684 & 25.467 & 9.14\% & 11.901 \\
			& Greedy & 4.606 & 0.39\% & 0.248 & 8.403 & 17.19\% & 1.274 & 14.311 & 13.36\% & 3.514 & 25.471 & 9.22\% & 11.992 \\
			\hline
			\multirow{3}{*}{VRPL} 
			& Ours   & \textbf{5.000} & \textbf{1.80\%} & 0.257 & \textbf{9.396} & \textbf{37.50\%} & 1.046 & \textbf{16.240} & \textbf{82.73\%} & 3.052 & \textbf{29.504} & \textbf{90.62\%} & 12.194 \\
			& Random & 5.002 & 0.31\% & 0.242 & 9.422 & 15.78\% & 1.262 & 16.422 & 7.97\% & 3.551 & 29.877 & 4.61\% & 12.338 \\
			& Greedy & 5.002 & 0.39\% & 0.245 & 9.420 & 14.14\% & 1.274 & 16.420 & 8.67\% & 3.591 & 29.880 & 4.45\% & 12.463 \\
			\hline
			\multirow{3}{*}{OVRP} 
			& Ours   & 3.484 & \textbf{0.39\%} & 0.245 & \textbf{6.160} & \textbf{19.84\%} & 0.950 & \textbf{10.105} & \textbf{46.41\%} & 2.761 & \textbf{17.555} & \textbf{51.41\%} & 10.926 \\
			& Random & 3.484 & 0.08\% & 0.230 & 6.166 & 13.20\% & 1.116 & 10.139 & 26.72\% & 3.198 & 17.622 & 22.73\% & 11.033 \\
			& Greedy & 3.484 & 0.16\% & 0.236 & 6.166 & 15.39\% & 1.129 & 10.137 & 24.30\% & 3.238 & 17.618 & 24.22\% & 11.245 \\
			\hline
			\multirow{3}{*}{VRPTW} 
			& Ours   & \textbf{7.657} & \textbf{7.27\%} & 0.295 & \textbf{14.728} & \textbf{28.12\%} & 1.439 & \textbf{25.174} & \textbf{36.88\%} & 3.974 & 44.946 & \textbf{37.03\%} & 14.438 \\
			& Random & 7.662 & 1.56\% & 0.319 & 14.746 & 20.08\% & 1.892 & 25.198 & 26.48\% & 4.922 & 44.926 & 28.12\% & 15.014 \\
			& Greedy & 7.664 & 1.41\% & 0.322 & 14.745 & 20.47\% & 1.904 & 25.198 & 28.20\% & 4.945 & \textbf{44.911} & 30.86\% & 15.308 \\
			\hline
			\multirow{3}{*}{TSP} 
			& Ours   & 3.824 & 0.00\% & 0.233 & \textbf{5.688} & \textbf{28.59\%} & 0.809 & \textbf{7.780} & \textbf{96.64\%} & 2.379 & \textbf{10.900} & \textbf{99.61\%} & 9.232 \\
			& Random & 3.824 & 0.00\% & 0.207 & 5.716 & 1.09\% & 0.947 & 8.019 & 1.25\% & 2.780 & 11.528 & 0.00\% & 9.820 \\
			& Greedy & 3.824 & 0.00\% & 0.204 & 5.715 & 2.19\% & 0.951 & 8.003 & 1.48\% & 2.769 & 11.395 & 0.39\% & 9.804 \\
			\hline
			\multirow{3}{*}{Average} 
			& Ours   & \textbf{4.865} & \textbf{2.29\%} & 0.260 & \textbf{8.754} & \textbf{31.11\%} & 1.050 & \textbf{14.456} & \textbf{70.59\%} & 2.997 & \textbf{25.313} & \textbf{75.87\%} & 11.481 \\
			& Random & 4.871 & 0.46\% & 0.251 & 8.775 & 14.19\% & 1.281 & 14.592 & 13.84\% & 3.582 & 25.600 & 10.77\% & 11.988 \\
			& Greedy & 4.868 & 0.57\% & 0.251 & 8.774 & 14.38\% & 1.294 & 14.590 & 13.53\% & 3.578 & 25.575 & 11.93\% & 12.121 \\
			\hline
		\end{tabular}
	}
	}
\end{table*}

\subsection{Impact of integrating the E-GAT with POMO}
\label{sec:5.5}

Another key contribution of our work is the integration of POMO \citep{Kwon2020} with the E-GAT encoder \citep{Lei2022}. The residual E-GAT, originally built on top of the attention model \citep{Kool2019}, aims at improving the learning process by incorporating edge information of all nodes in a problem. This allows the model to capture richer contextual information, resulting in more accurate attention coefficient computations. When combined with POMO, this improved encoding can lead to a better utilization of POMO’s multiple starting nodes, generating higher quality solutions.

\begin{figure*}[!ht]
	\begin{center}
		\setlength{\unitlength}{\textwidth}
		\begin{picture}(1,0.25) 
			\put(0.17,0.22){(a)}
			\put(0.50,0.22){(b)}
			\put(0.84,0.22){(c)}
			\includegraphics[width=\textwidth]{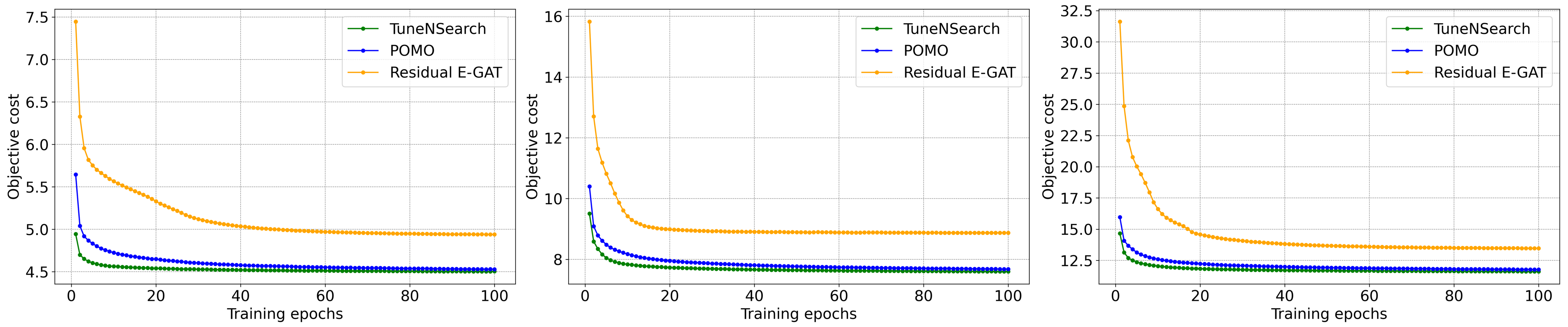}
		\end{picture}
		\caption{MDVRP training curves of POMO, Residual E-GAT, and TuneNSearch - (a) $n$ = 20; (b) $n$ = 50; (c) $n$ = 100.}
		\label{fig6}
	\end{center}
\end{figure*}

In this subsection, we compare the performance of TuneNSearch with both the original POMO and residual E-GAT models. Our aim is to analyze how effective the encoding capability of each approach is, and how well they capture the graph features of the VRP. We begin by examining the training patterns of all three approaches, analyzing the evolution of the average objective function at each epoch, illustrated in Fig.~\ref{fig6}. The models were trained in MDVRP instances with 20, 50 and 100 customers and 2, 3 and 4 depots, respectively. Across all three problem sizes, TuneNSearch exhibited a more efficient convergence. In comparison to the residual E-GAT model, the difference was a lot more noticeable, since it does not incorporate the exploitation of solutions symmetries and multiple starting nodes introduced by POMO. 

In Table~\ref{table10}, we compared the inference results of all three methods. The evaluation was performed on 1280 randomly generated instances, using a greedy decoding. Both POMO and TuneNSearch utilized instance augmentation, with no local search performed after inference. We note that the residual E-GAT model did not use instance augmentation, as it does not exploit solution symmetries. 

The results show that TuneNSearch outperforms the other methods in all problem sizes. These findings confirm that integrating the residual E-GAT with POMO improves learning performance, enabling a more effective encoding of the problem’s features. 

\begin{table}[!ht]
	\centering
	\caption{Average inference results on 1280 randomly generated MDVRP instances: impact of integrating E-GAT with POMO.} \label{table10}
	\normalsize
	\begin{tabular}{c|cc|cc|cc}
		\hline
		\multirow{2}{*}{Method} & \multicolumn{2}{c|}{n = 20} & \multicolumn{2}{c|}{n = 50} & \multicolumn{2}{c}{n = 100} \\
		& Obj. & Time (m) & Obj. & Time (m) & Obj. & Time (m) \\
		\hline
		Ours & \textbf{4.526} & 0.052 & \textbf{7.647} & 0.135 & \textbf{11.734} & 0.501 \\
		POMO & 4.537 & 0.040 & 7.691 & 0.084 & 11.840 & 0.291 \\
		Residual E-GAT & 4.941 & 0.065 & 10.585 & 0.080 & 13.130 & 0.116 \\
		\hline
	\end{tabular}
\end{table}

\subsection{Analysis of the fine-tuning phase}
\label{sec:5.6}

\begin{figure*}[!ht]
	\begin{center}
		\setlength{\unitlength}{\textwidth}
		\begin{picture}(1,0.25) 
			\put(0.16,0.22){(a)}
			\put(0.50,0.22){(b)}
			\put(0.83,0.22){(c)}
			\includegraphics[width=\textwidth]{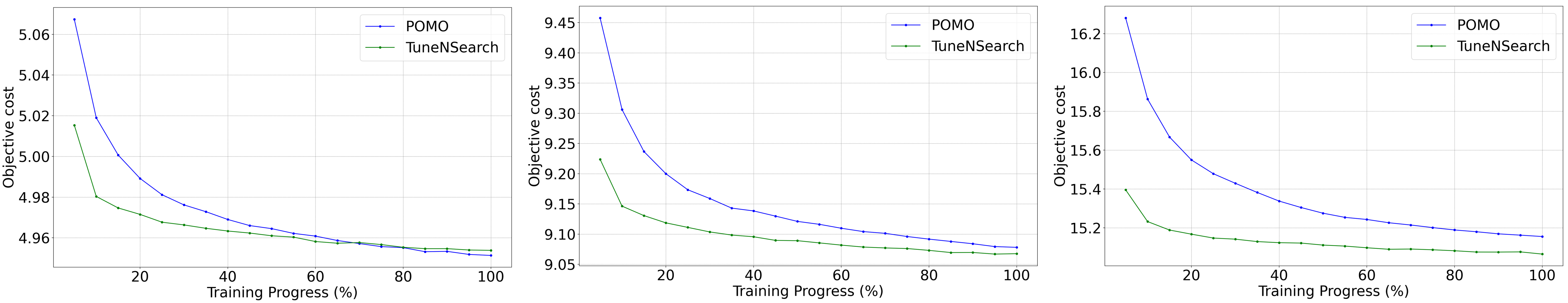}
		\end{picture}
		\caption{Training curves of POMO and TuneNSearch - (a) $n$ = 20; (b) $n$ = 50; (c) $n$ = 100.}
		\label{fig7}
	\end{center}
\end{figure*}

Besides evaluating the performance during inference, we also compare the training patterns of POMO and TuneNSearch (only the fine-tuned models, excluding the MDVRP), averaged across all VRP variants (see Fig.~\ref{fig7}). Since POMO was trained for 100 epochs while TuneNSearch underwent fine-tuning for 20 epochs, we normalized the training progress, presenting it as a percentage rather than using the number of epochs.

For instances with 20 customers, POMO slightly outperforms TuneNSearch towards the end of training, however, for problems with 50 and 100 customers, POMO is inferior to TuneNSearch. Notably, the efficiency of our method is particularly evident at the beginning of training, since it benefits from prior knowledge gained during the pre-training phase. Furthermore, the performance difference between TuneNSearch and POMO appears to widen as problem sizes increase.

Moreover, Table~\ref{table11} presents the average training times for all models considered in this study. We note that the results of POMO and TuneNSearch are aggregated across all VRP variants. As shown, TuneNSearch requires about one-third of the computational time needed to train POMO from the beginning for each variant. Nonetheless, the strength of TuneNSearch lies in its capability to solve multiple VRP variants with the same pre-trained model and a fast fine-tuning phase.

\begin{table}[!ht]
	\centering
	\caption{Average training time (m) for all models.} \label{table11}
	\normalsize
	\begin{tabular}{c|c|c|c}
		\hline
		& $n$ = 20 & $n$ = 50 & $n$ = 100 \\
		\hline
		MD-MTA & 507.52 & 1690.24 & 3525.12 \\
		POMO & 144.88 & 578.29 & 1280.45 \\
		Ours (pre-trained MDVRP) & 243.07 & 1119.33 & 2207.62 \\
		Ours (fine-tuning) & 47.96 & 192.08 & 391.91 \\
		\hline
	\end{tabular}
\end{table}

\subsection{Computational runtime complexity}
\label{sec:5.7}

\textcolor{blue}{Next, we investigate the computational runtime complexity of TuneNSearch on a set of benchmark instances, which are significantly large, containing up to 1000 nodes.} The primary goal of this analysis is to understand how the integration of the local search algorithm scales with increasing problem sizes.

\begin{figure*}[!ht]
	\begin{center}
		\includegraphics[width=\textwidth]{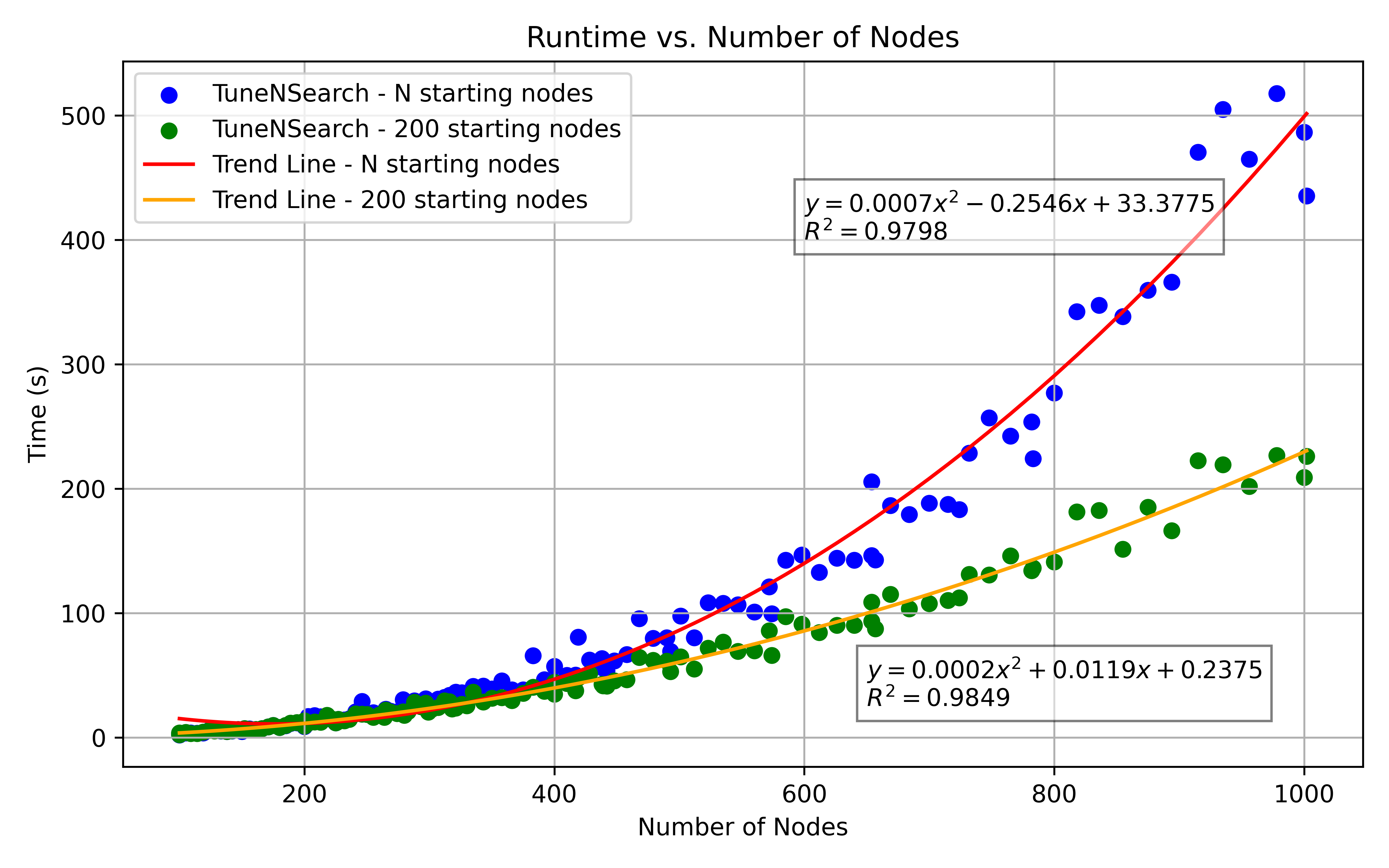}
		\caption{TuneNSearch computational runtime on varying instance sizes.} \label{fig8}
	\end{center}
\end{figure*}

The experimental results, presented in Fig.~\ref{fig8}, reveal that TuneNSearch exhibits polynomial runtime growth with respect to instance size (defined as a function of the number of nodes). More specifically, the trend follows a quadratic pattern, as indicated by the fitted curve equation of $y = 0.0007x^2 - 0.2546x + 33.3775$, with an $R^2$ value of 0.9798.  While this scaling behavior is encouraging, especially compared to traditional methods, which typically present an exponential complexity, there is still room for improvement in computational efficiency. One such opportunity lies in reducing the number of starting nodes used by the decoder. Our decoder generates multiple candidate solutions per instance by initializing from every possible node. Although this approach may improve solution quality, it increases the computational runtime. To address this, we reevaluated TuneNSearch on all benchmark problems while limiting the maximum number of starting nodes to 200. The results show that this adjustment leads to a substantial reduction in runtime, particularly for larger instances. With this modification, the runtime still follows a polynomial trend, but with a more favorable curve equation of $y = 0.0002^2 + 0.0119x + 0.2375$, and an improved $R^2$ value of 0.9849. Furthermore, reducing the number of starting nodes to 200 had a negligible impact on the algorithm's effectiveness, with the average solution quality deteriorating by less than 1\%.

\section{Conclusion}
\label{sec:conclusion}

In this paper, we introduce TuneNSearch, a novel transfer learning method designed for adaptation to various VRP variants through an efficient fine-tuning phase. Our approach builds on the model architecture of \citet{Kwon2021} by enhancing the encoder with an edge-aware mechanism. While \citet{Lei2022} previously integrated this technique into the attention model \citep{Kool2019}, we extend its application to POMO, demonstrating that this extension improves the model's ability to encode the problem’s features more effectively. \textcolor{blue}{We first pre-train our model using MDVRP data, exploiting its complex graph-structured features.} This strategy allows for a broader generalization across a variety of VRP variants, including both single- and multi-depot tasks. To evaluate the effectiveness of our approach, we compared it to an identical model pre-trained on the CVRP. The results demonstrate that while both methods perform similarly on single-depot variants, TuneNSearch substantially outperforms the CVRP-based model on multi-depot variants. To validate the learning process on the MDVRP, we compared TuneNSearch with MD-MTA, demonstrating superior results across all problem sizes while requiring much less training time. Finally, we integrated an efficient local search method to refine the quality of solutions generated by our model, leading to significant performance improvements with a small computational overhead. We also conducted experiments to assess the impact of increasing the number of local search iterations. Our findings indicate that more constrained tasks tend to benefit more from this procedure.

To evaluate the generalization of TuneNSearch, we performed extensive experiments on numerous VRP variants and baselines. Experimental results on randomly generated instances show that TuneNSearch outperforms POMO, which is specialized for each variant, while requiring only one-fifth of the total training epochs. Moreover, TuneNSearch achieves performance comparable to OR-Tools guided local search procedure, often even outperforming it on many occasions, delivering results at a fraction of the computational time. We also provide results on benchmark instances of different VRP variants, where TuneNSearch outperforms other state-of-the-art neural-based models on most problems, consistently achieving higher-quality solutions. These findings demonstrate not only TuneNSearch’s strong cross-task generalization, but also its cross-size and cross-distribution generalization.

\subsection{Limitations and future work}
\label{sec:6.1}

While this study offers insights into the development of generalizable neural-based methods, there are certain limitations to our approach. First, TuneNSearch is specifically designed to solve traditional VRPs, with the constraints outlined in Section~\ref{sec:3.3} (or any combination of such constraints). Although it is possible to extend our method to other problems, such as the orienteering problem or prize collecting TSP, doing so would require manual adjustments to the Transformer architecture to accommodate new constraints. \textcolor{blue}{Second, TuneNSearch assumes that the objective function (minimizing the total distance traveled) remains unchanged across all scenarios.} Modifying the objective during the fine-tuning phase may impact the learning process, as the neurons of the pre-trained model are already optimized for distance minimization.

Looking ahead, we aim to extend TuneNSearch beyond traditional VRP variants to address a broader range of combinatorial problems. In particular, we plan to adapt our method to scheduling problems, which can be framed as variations of the TSP. Another promising direction is to further enhance the performance of TuneNSearch to improve solution quality even further. One possible approach could involve clustering training instances based on their underlying distributions and training specialized local models for each subset, which could potentially improve generalization. Alternatively, integrating decomposition approaches with learning-based methods to solve the VRP in a “divide-and-conquer” manner is another promising research direction.

\section*{Acknowledgements}

This work has been supported by the European Union under the Next Generation EU, through a grant of the Portuguese Republic's Recovery and Resilience Plan Partnership Agreement [project C645808870-00000067], within the scope of the project PRODUTECH R3 – "Agenda Mobilizadora da Fileira das Tecnologias de Produção para a Reindustrialização", Total project investment: 166.988.013,71 Euros; Total Grant: 97.111.730,27 Euros; and by the European Regional Development Fund (ERDF) through the Operational Program for Competitiveness and Internationalization (COMPETE 2020) under the project POCI-01-0247-FEDER-046102 (PRODUTECH4S\&C); and by national funds through FCT – Fundação para a Ciência e a Tecnologia, under projects UID/00285 - Centre for Mechanical Engineering, Materials and Processes and LA/P/0112/2020.

\bibliographystyle{elsarticle-harv} 
\bibliography{./elsarticle-harv}

\appendix
\setcounter{figure}{0}
\setcounter{table}{0}

\section{Hyper-parameters sensitivity}
\label{sec:appA}

In this appendix, we analyze the sensitivities of various hyper-parameters in our approach. Specifically, we examine the impact of the number of encoder layers $L$, the number of heads in the decoder $H$, the hidden dimension $h_x$ and the hidden edge dimension $h_e$.

\begin{figure}[!h]
	\begin{center}
		\includegraphics[width=0.9\textwidth]{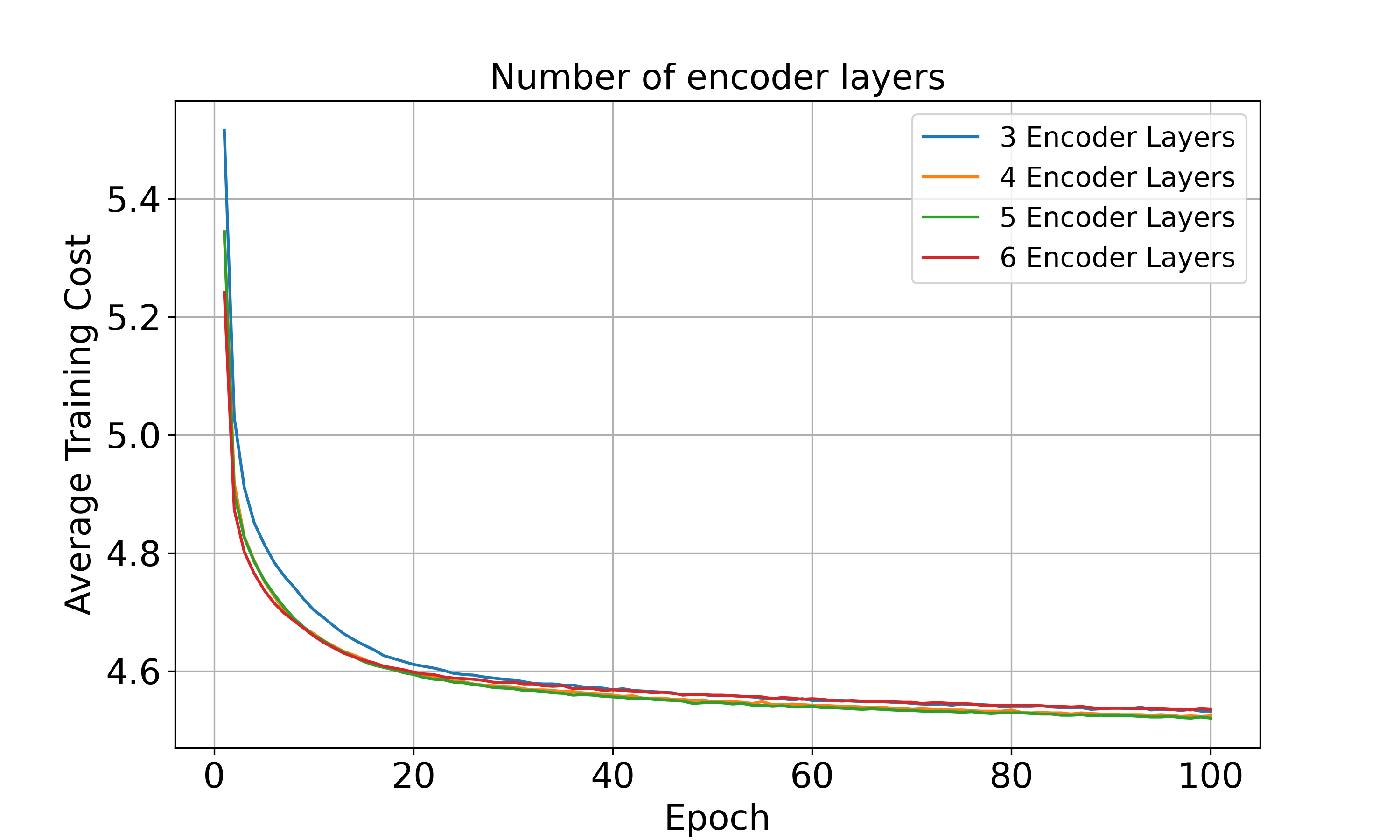}
		\caption{Sensitivity analysis of the number of encoder layers $L$.} \label{figA1}
	\end{center}
\end{figure}

First, we assessed the model's sensitivity to the number of encoder layers by testing values of $L = {3,4,5,6}$. Fig.~\ref{figA1} presents the average cost during pre-training for each value over 100 episodes. Interestingly, the model's performance degraded when using 6 encoder layers, yielding results comparable to those obtained with 3 layers. In contrast, models with 4 and 5 encoder layers performed better, with the 5-layer configuration showing a slight edge over the 4-layer model.

\begin{figure}[!h]
	\begin{center}
		\includegraphics[width=0.9\textwidth]{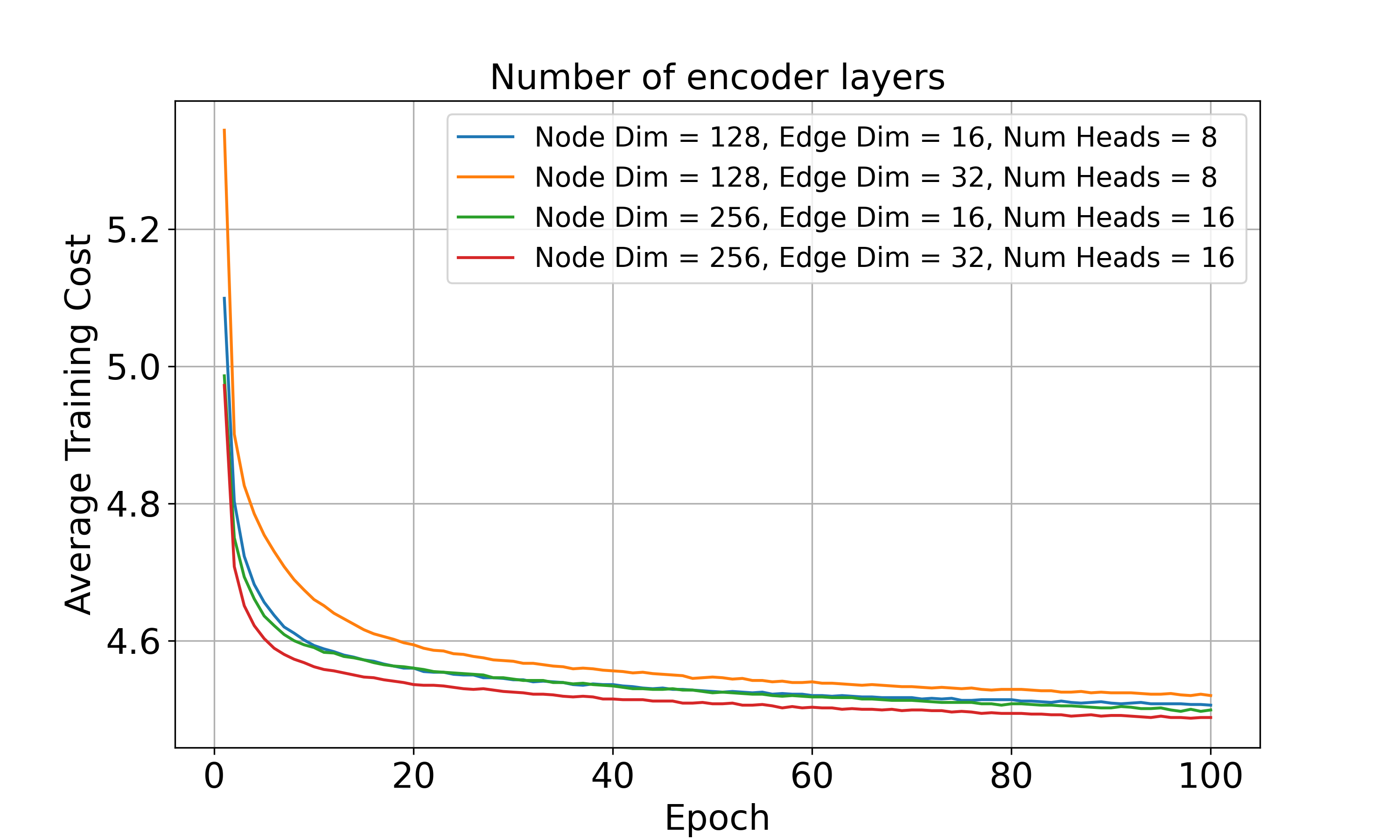}
		\caption{Sensitivity analysis of the hidden dimension $h_x$, hidden edge dimension $h_e$ and number of heads $H$.} \label{figA2}
	\end{center}
\end{figure}

For the other hyper-parameters, we explored four different combinations, as shown in Fig.~\ref{figA2}. The best-performing configuration used a hidden dimension of $h_x$  = 256, a hidden edge dimension of $h_e$  = 32 and $H$ = 16 heads in the decoder.

\section{Instance generation}
\label{sec:appB}

To generate the random training and testing instances for all VRP variants (including $n$ = 20, $n$ = 50 and $n$ = 100), we uniformly sample node coordinates within a [0, 1] x [0, 1] Euclidean space. Customer demands are randomly sampled from {1, ..., 9} and then normalized with respect to vehicle capacity, which is set to 50. Additional data is generated depending on the VRP variant considered. We follow similar settings as other research, which are described next. 1) VRPB: We randomly select 20\% of the customers as backhauls, following \citet{Liu2024}. Our paper considers mixed backhauls, meaning linehaul and backhaul customers can be visited without a strict precedence order. However, the vehicle’s capacity constraints must always be respected. For this reason, every route must start with a linehaul customer. 2) VRPL: We set a length limit of 3 for all routes, again following \citet{Liu2024}. 3) VRPTW: For the VRPTW, we follow the same procedure as \citet{Solomon1987}, \citet{Li2021} and \citet{Zhou2024} for generating the service times and time windows.

\section{Generalization on INCOM2024 benchmark instances}
\label{sec:appC}

We present results on additional MDVRP benchmark instances on Table~\ref{tableC1}. We evaluated TuneNSearch performance using instances from INCOM 2024 data-drive logistics challenge dataset \citep{Xu2024}. This dataset was introduced by the Supply Chain AI Lab (SCAIL) of the University of Cambridge at the INCOM 2024 Conference for a logistics challenge, which can be downloaded via the link \footnote{https://www.ifm.eng.cam.ac.uk/research/supply-chain-ai-lab/data-competition/}. It includes 100 instances with problem sizes ranging from 100 to 1000 nodes. To benchmark this dataset, we executed PyVRP using the same time limit as \citet{Vidal2022} and \citet{Wouda2024}. We used the same experimental setup from Section~\ref{sec:5.2}. As with Cordeau’s dataset, TuneNSearch consistently outperformed the MD-MTA and POMO models. These results further demonstrate the strong generalization capabilities of TuneNSearch, even when applied to problems of larger scale with different distributions.

\setlength{\tabcolsep}{4pt} 
\renewcommand{\arraystretch}{1.1}
\small
\begin{longtable}[!h]{cccc|cccccc}
	\caption{Generalization on INCOM 2024 benchmark instances.} \label{tableC1} \\
	\hline
	\multirow{2}{*}{Instance} & \multirow{2}{*}{Depots} & \multirow{2}{*}{Customers} & \multirow{2}{*}{HGS-PyVRP} & \multicolumn{2}{c}{Ours} & \multicolumn{2}{c}{MD-MTA} & \multicolumn{2}{c}{POMO} \\
	&  &  &  & Obj. & RPD & Obj. & RPD & Obj. & RPD \\
	\hline
	scail01 & 4 & 100 & 13099 & \textbf{13315} & \textbf{1.649   \%} & 13831 & 5.588 \% & 13855 & 5.771   \% \\
	scail02 & 3 & 105 & 9590 & \textbf{9727} & \textbf{1.429   \%} & 10023 & 4.515 \% & 10153 & 5.871   \% \\
	scail03 & 3 & 110 & 11772 & \textbf{11929} & \textbf{1.334   \%} & 12273 & 4.256 \% & 12257 & 4.120   \% \\
	scail04 & 4 & 115 & 10600 & \textbf{10714} & \textbf{1.075   \%} & 11182 & 5.491 \% & 10987 & 3.651   \% \\
	scail05 & 2 & 119 & 16374 & \textbf{16739} & \textbf{2.229   \%} & 17196 & 5.020 \% & 16965 & 3.609   \% \\
	scail06 & 4 & 123 & 11159 & \textbf{11329} & \textbf{1.523   \%} & 11934 & 6.945 \% & 11842 & 6.121   \% \\
	scail07 & 3 & 127 & 8544 & \textbf{8690} & \textbf{1.709   \%} & 10160 & 18.914 \% & 9573 & 12.044   \% \\
	scail08 & 2 & 132 & 11846 & \textbf{12011} & \textbf{1.393   \%} & 12502 & 5.538 \% & 12229 & 3.233 \% \\
	scail09 & 4 & 137 & 8811 & \textbf{9091} & \textbf{3.178   \%} & 10223 & 16.025 \% & 9840 & 11.679   \% \\
	scail10 & 2 & 142 & 17173 & \textbf{17196} & \textbf{0.134   \%} & 18101 & 5.404 \% & 18088 & 5.328   \% \\
	scail11 & 3 & 147 & 10626 & \textbf{10934} & \textbf{2.899   \%} & 12010 & 13.025 \% & 11604 & 9.204   \% \\
	scail12 & 4 & 151 & 13504 & \textbf{13800} & \textbf{2.192   \%} & 14498 & 7.361 \% & 14831 & 9.827   \% \\
	scail13 & 3 & 156 & 13312 & \textbf{13578} & \textbf{1.998   \%} & 14576 & 9.495 \% & 15522 & 16.602   \% \\
	scail14 & 2 & 161 & 9785 & \textbf{9984} & \textbf{2.034   \%} & 10735 & 9.709 \% & 10756 & 9.923   \% \\
	scail15 & 4 & 165 & 16071 & \textbf{16690} & \textbf{3.852   \%} & 17917 & 11.486 \% & 17631 & 9.707   \% \\
	scail16 & 3 & 170 & 15011 & \textbf{15205} & \textbf{1.292   \%} & 16149 & 7.581 \% & 16255 & 8.287   \% \\
	scail17 & 3 & 175 & 13712 & \textbf{13937} & \textbf{1.641   \%} & 14925 & 8.846 \% & 14616 & 6.593   \% \\
	scail18 & 2 & 180 & 15144 & \textbf{15384} & \textbf{1.585   \%} & 16469 & 8.749 \% & 16514 & 9.046   \% \\
	scail19 & 2 & 184 & 20116 & \textbf{20710} & \textbf{2.953   \%} & 21518 & 6.970 \% & 21615 & 7.452   \% \\
	scail20 & 2 & 189 & 24494 & \textbf{25012} & \textbf{2.115   \%} & 26196 & 6.949 \% & 26504 & 8.206   \% \\
	scail21 & 2 & 194 & 14079 & \textbf{14415} & \textbf{2.387   \%} & 15592 & 10.746 \% & 16350 & 16.130   \% \\
	scail22 & 4 & 198 & 35731 & \textbf{35750} & \textbf{0.053   \%} & 38619 & 8.083 \% & 38663 & 8.206   \% \\
	scail23 & 4 & 202 & 13357 & \textbf{13857} & \textbf{3.743   \%} & 14996 & 12.271 \% & 14753 & 10.451   \% \\
	scail24 & 2 & 206 & 37615 & \textbf{38417} & \textbf{2.132   \%} & 40864 & 8.637 \% & 41506 & 10.344   \% \\
	scail25 & 3 & 210 & 19067 & \textbf{19735} & \textbf{3.503   \%} & 20680 & 8.460 \% & 20888 & 9.551   \% \\
	scail26 & 3 & 214 & 12428 & \textbf{12749} & \textbf{2.583   \%} & 13840 & 11.361 \% & 14154 & 13.888   \% \\
	scail27 & 4 & 219 & 21051 & \textbf{21748} & \textbf{3.311   \%} & 23309 & 10.726 \% & 23643 & 12.313   \% \\
	scail28 & 3 & 224 & 13236 & \textbf{13796} & \textbf{4.231   \%} & 14574 & 10.109 \% & 14860 & 12.270   \% \\
	scail29 & 3 & 228 & 29825 & \textbf{30508} & \textbf{2.290   \%} & 32370 & 8.533 \% & 33266 & 11.537   \% \\
	scail30 & 3 & 233 & 33578 & \textbf{34537} & \textbf{2.856   \%} & 37066 & 10.388 \% & 36983 & 10.141   \% \\
	scail31 & 2 & 238 & 31881 & \textbf{32297} & \textbf{1.305   \%} & 35048 & 9.933 \% & 34649 & 8.682   \% \\
	scail32 & 3 & 242 & 11293 & \textbf{11703} & \textbf{3.631   \%} & 13337 & 18.100 \% & 13974 & 23.740   \% \\
	scail33 & 2 & 247 & 27116 & \textbf{27907} & \textbf{2.917   \%} & 29881 & 10.197 \% & 31910 & 17.680   \% \\
	scail34 & 4 & 252 & 20586 & \textbf{20924} & \textbf{1.642   \%} & 22623 & 9.895 \% & 22777 & 10.643   \% \\
	scail35 & 2 & 257 & 43163 & \textbf{44465} & \textbf{3.016   \%} & 47174 & 9.293 \% & 47704 & 10.521   \% \\
	scail36 & 2 & 262 & 41070 & \textbf{42635} & \textbf{3.811   \%} & 44628 & 8.663 \% & 45390 & 10.519   \% \\
	scail37 & 3 & 266 & 17771 & \textbf{17959} & \textbf{1.058   \%} & 19691 & 10.804 \% & 20176 & 13.533   \% \\
	scail38 & 4 & 270 & 20579 & \textbf{21406} & \textbf{4.019   \%} & 23223 & 12.848 \% & 23172 & 12.600   \% \\
	scail39 & 3 & 275 & 27075 & \textbf{27390} & \textbf{1.163   \%} & 29875 & 10.342 \% & 30323 & 11.996   \% \\
	scail40 & 2 & 279 & 28127 & \textbf{29162} & \textbf{3.680   \%} & 31097 & 10.559 \% & 31583 & 12.287   \% \\
	scail41 & 3 & 284 & 26304 & \textbf{27263} & \textbf{3.646   \%} & 30426 & 15.671 \% & 30467 & 15.826   \% \\
	scail42 & 3 & 288 & 17468 & \textbf{18343} & \textbf{5.009   \%} & 19965 & 14.295 \% & 20089 & 15.005   \% \\
	scail43 & 2 & 293 & 27788 & \textbf{28379} & \textbf{2.127   \%} & 30663 & 10.346 \% & 32270 & 16.129   \% \\
	scail44 & 3 & 297 & 25846 & \textbf{26859} & \textbf{3.919   \%} & 29160 & 12.822 \% & 29033 & 12.331   \% \\
	scail45 & 2 & 302 & 27694 & \textbf{28639} & \textbf{3.412   \%} & 31177 & 12.577 \% & 31445 & 13.544   \% \\
	scail46 & 3 & 309 & 35469 & \textbf{36498} & \textbf{2.901   \%} & 39411 & 11.114 \% & 40518 & 14.235   \% \\
	scail47 & 2 & 315 & 16346 & \textbf{17035} & \textbf{4.215   \%} & 18443 & 12.829 \% & 19669 & 20.329   \% \\
	scail48 & 2 & 323 & 24758 & \textbf{25840} & \textbf{4.370   \%} & 28175 & 13.802 \% & 27146 & 9.645   \% \\
	scail49 & 3 & 330 & 22161 & \textbf{22507} & \textbf{1.561   \%} & 25948 & 17.089 \% & 28425 & 28.266   \% \\
	scail50 & 3 & 337 & 13857 & \textbf{14644} & \textbf{5.679   \%} & 16729 & 20.726 \% & 17235 & 24.378   \% \\
	scail51 & 3 & 345 & 24642 & \textbf{25799} & \textbf{4.695   \%} & 28033 & 13.761 \% & 28790 & 16.833   \% \\
	scail52 & 3 & 352 & 17769 & \textbf{18774} & \textbf{5.656   \%} & 20833 & 17.243 \% & 21473 & 20.845   \% \\
	scail53 & 3 & 358 & 25662 & \textbf{26526} & \textbf{3.367   \%} & 28809 & 12.263 \% & 29691 & 15.700   \% \\
	scail54 & 2 & 364 & 32397 & \textbf{33196} & \textbf{2.466   \%} & 36680 & 13.220 \% & 37105 & 14.532   \% \\
	scail55 & 3 & 370 & 21246 & \textbf{22448} & \textbf{5.658   \%} & 24650 & 16.022 \% & 25102 & 18.149   \% \\
	scail56 & 3 & 377 & 18647 & \textbf{19261} & \textbf{3.293   \%} & 21757 & 16.678 \% & 22436 & 20.320   \% \\
	scail57 & 2 & 385 & 38457 & \textbf{40063} & \textbf{4.176   \%} & 43683 & 13.589 \% & 44039 & 14.515   \% \\
	scail58 & 3 & 393 & 17678 & \textbf{18773} & \textbf{6.194   \%} & 21156 & 19.674 \% & 22078 & 24.890   \% \\
	scail59 & 3 & 401 & 86457 & \textbf{87953} & \textbf{1.730   \%} & 96866 & 12.039 \% & 103530 & 19.747   \% \\
	scail60 & 4 & 413 & 30626 & \textbf{32282} & \textbf{5.407   \%} & 35540 & 16.045 \% & 36085 & 17.825   \% \\
	scail61 & 3 & 424 & 33534 & \textbf{34993} & \textbf{4.351   \%} & 39134 & 16.699 \% & 39060 & 16.479   \% \\
	scail62 & 3 & 438 & 29444 & \textbf{30866} & \textbf{4.830   \%} & 33676 & 14.373 \% & 35219 & 19.613   \% \\
	scail63 & 2 & 452 & 33495 & \textbf{33554} & \textbf{0.176   \%} & 37117 & 10.813 \% & 38676 & 15.468   \% \\
	scail64 & 3 & 466 & 31934 & \textbf{33371} & \textbf{4.500   \%} & 37029 & 15.955 \% & 37896 & 18.670   \% \\
	scail65 & 3 & 480 & 17811 & \textbf{18780} & \textbf{5.440   \%} & 21959 & 23.289 \% & 23293 & 30.779   \% \\
	scail66 & 3 & 493 & 68424 & \textbf{69353} & \textbf{1.358   \%} & 75055 & 9.691 \% & 79200 & 15.749   \% \\
	scail67 & 4 & 506 & 41279 & \textbf{42795} & \textbf{3.673   \%} & 47398 & 14.823 \% & 49355 & 19.564   \% \\
	scail68 & 4 & 516 & 30812 & \textbf{32249} & \textbf{4.664   \%} & 36866 & 19.648 \% & 36288 & 17.772   \% \\
	scail69 & 3 & 526 & 23097 & \textbf{23984} & \textbf{3.840   \%} & 34706 & 50.262 \% & 37294 & 61.467   \% \\
	scail70 & 2 & 535 & 30467 & \textbf{31229} & \textbf{2.501   \%} & 36158 & 18.679 \% & 39932 & 31.066   \% \\
	scail71 & 3 & 548 & 44511 & \textbf{45454} & \textbf{2.119   \%} & 51096 & 14.794 \% & 52273 & 17.438   \% \\
	scail72 & 2 & 560 & 58858 & \textbf{60956} & \textbf{3.565   \%} & 67969 & 15.480 \% & 69224 & 17.612   \% \\
	scail73 & 4 & 573 & 26192 & \textbf{27230} & \textbf{3.963   \%} & 31052 & 18.555 \% & 34598 & 32.094   \% \\
	scail74 & 2 & 587 & 63596 & \textbf{66061} & \textbf{3.876   \%} & 72128 & 13.416 \% & 73651 & 15.811   \% \\
	scail75 & 2 & 597 & 55622 & \textbf{57610} & \textbf{3.574   \%} & 66019 & 18.692 \% & 67546 & 21.438   \% \\
	scail76 & 3 & 609 & 41069 & \textbf{42303} & \textbf{3.005   \%} & 47966 & 16.794 \% & 53417 & 30.066   \% \\
	scail77 & 3 & 625 & 102859 & \textbf{104795} & \textbf{1.882   \%} & 117132 & 13.876 \% & 119924 & 16.591   \% \\
	scail78 & 3 & 641 & 43495 & \textbf{45432} & \textbf{4.453   \%} & 50906 & 17.039 \% & 53488 & 22.975   \% \\
	scail79 & 4 & 656 & 42247 & \textbf{43888} & \textbf{3.884   \%} & 50346 & 19.171 \% & 53284 & 26.125   \% \\
	scail80 & 4 & 672 & 46051 & \textbf{47185} & \textbf{2.462   \%} & 53573 & 16.334 \% & 58291 & 26.579   \% \\
	scail81 & 3 & 687 & 21921 & \textbf{23790} & \textbf{8.526   \%} & 28812 & 31.436 \% & 30874 & 40.842   \% \\
	scail82 & 4 & 702 & 49390 & \textbf{50789} & \textbf{2.833   \%} & 57085 & 15.580 \% & 66046 & 33.723   \% \\
	scail83 & 3 & 717 & 49456 & \textbf{51915} & \textbf{4.972   \%} & 58367 & 18.018 \% & 62634 & 26.646   \% \\
	scail84 & 4 & 733 & 18124 & \textbf{19757} & \textbf{9.010   \%} & 24508 & 35.224 \% & 28168 & 55.418   \% \\
	scail85 & 2 & 748 & 51616 & \textbf{54245} & \textbf{5.093   \%} & 61848 & 19.823 \% & 62601 & 21.282   \% \\
	scail86 & 3 & 763 & 33514 & \textbf{35387} & \textbf{5.589   \%} & 40891 & 22.012 \% & 43619 & 30.152   \% \\
	scail87 & 3 & 779 & 31036 & \textbf{33101} & \textbf{6.654   \%} & 39041 & 25.793 \% & 45531 & 46.704   \% \\
	scail88 & 3 & 795 & 41243 & \textbf{43654} & \textbf{5.846   \%} & 53080 & 28.701 \% & 54856 & 33.007   \% \\
	scail89 & 4 & 811 & 22990 & \textbf{24662} & \textbf{7.273   \%} & 31070 & 35.146 \% & 34960 & 52.066   \% \\
	scail90 & 3 & 826 & 68092 & \textbf{69168} & \textbf{1.580   \%} & 77239 & 13.433 \% & 82490 & 21.145   \% \\
	scail91 & 4 & 841 & 22835 & \textbf{24227} & \textbf{6.096   \%} & 31508 & 37.981 \% & 38902 & 70.361   \% \\
	scail92 & 4 & 856 & 58744 & \textbf{59867} & \textbf{1.912   \%} & 71191 & 21.188 \% & 74127 & 26.187   \% \\
	scail93 & 4 & 871 & 49517 & \textbf{50694} & \textbf{2.377   \%} & 59129 & 19.411 \% & 67289 & 35.891   \% \\
	scail94 & 2 & 887 & 76542 & \textbf{78704} & \textbf{2.825   \%} & 90743 & 18.553 \% & 96481 & 26.050   \% \\
	scail95 & 2 & 903 & 107324 & \textbf{108076} & \textbf{0.701   \%} & 119771 & 11.596 \% & 135159 & 25.935   \% \\
	scail96 & 2 & 922 & 82326 & \textbf{82908} & \textbf{0.707   \%} & 95933 & 16.528 \% & 100659 & 22.269   \% \\
	scail97 & 2 & 942 & 103103 & \textbf{107592} & \textbf{4.354   \%} & 118724 & 15.151 \% & 122495 & 18.808   \% \\
	scail98 & 4 & 962 & 37379 & \textbf{40669} & \textbf{8.802   \%} & 48431 & 29.567 \% & 52647 & 40.846   \% \\
	scail99 & 4 & 982 & 37400 & \textbf{39561} & \textbf{5.778   \%} & 47544 & 27.123 \% & 58895 & 57.473   \% \\
	scail100 & 4 & 1002 & 29043 & \textbf{30601} & \textbf{5.364   \%} & 39199 & 34.969 \% & 45220 & 55.700   \% \\
	\hline
	\multicolumn{4}{c}{Avg. RPD} & \multicolumn{2}{c}{\textbf{3.354 \%}} & \multicolumn{2}{c}{15.052 \%} & \multicolumn{2}{c}{19.702 \%} \\
	\hline
\end{longtable}
\normalsize

\end{document}